\pdfoutput=1

\documentclass[review]{elsarticle}

\makeatletter
\def\ps@pprintTitle{%
 \let\@oddhead\@empty
 \let\@evenhead\@empty
 \def\@oddfoot{}%
 \let\@evenfoot\@oddfoot}
\makeatother

\usepackage{hyperref}
\usepackage{lineno}
\modulolinenumbers[5]
\usepackage[utf8]{inputenc}
\usepackage{amsmath}
\usepackage{amsfonts}
\usepackage{amssymb}
\usepackage{multirow,booktabs}
\usepackage{mathabx}
\usepackage{changepage}
\usepackage{enumitem}
\usepackage{url}

\journal{Signal Processing: Image Communication}

\begin{document}

\begin{frontmatter}

\begin{itemize}[align=parleft, labelsep=1cm]

\item[\textbf{Citation}]{D. Temel and G. AlRegib, CSV: Image quality assessment based on color, structure, and visual system, Signal Processing: Image Communication, Volume 48, 2016, Pages 92-103, ISSN 0923-5965}

\item[\textbf{DOI}]{https://doi.org/10.1016/j.image.2016.08.008}

\item[\textbf{Review}]{Submitted 10 February 2016, Accepted 31 August 2016}

\item[\textbf{Code}]{https://ghassanalregib.com/publication-3mt-videos/}

\item[\textbf{Bib}]{@article\{Temel2016\_SPIC,\\
title = "CSV: Image quality assessment based on color, structure, and visual system",\\
journal = "Signal Processing: Image Communication",\\
volume = "48",\\
pages = "92 - 103",\\
year = "2016",\\
issn = "0923-5965",\\
doi = "https://doi.org/10.1016/j.image.2016.08.008",\\
url = "http://www.sciencedirect.com/science/article/pii/S0923596516301151",\\
author = "Dogancan Temel and Ghassan AlRegib" \} }

\item[\textbf{Contact}]{alregib@gatech.edu~~~~~~~\url{https://ghassanalregib.com/}\\dcantemel@gmail.com~~~~~~~\url{http://cantemel.com/}}
\end{itemize}
\thispagestyle{empty}
\newpage
\clearpage
\setcounter{page}{1}

\title{CSV: Image Quality Assessment Based on Color, Structure, and Visual System}

\author{Dogancan Temel}
\author{Ghassan AlRegib}
\address{Center for Signal and Information Processing}
\address{School of Electrical and Computer Engineering}
\address{Georgia Institute of Technology, Atlanta, GA 30332}
\address{dcantemel@gmail.com,alregib@gatech.edu}

\begin{abstract}
This paper presents a full-reference image quality estimator based on \textbf{c}olor, \textbf{s}tructure, and \textbf{v}isual system characteristics denoted as \textbf{CSV}. In contrast to the majority of existing methods, we quantify perceptual color degradations rather than absolute pixel-wise changes. We use the CIEDE2000 color difference formulation to quantify low-level color degradations and the Earth  Mover's Distance between color name descriptors to measure significant color degradations. In addition to the perceptual color difference, \textbf{CSV} also contains structural and perceptual differences. Structural feature maps are obtained by mean subtraction and divisive normalization, and perceptual feature maps are obtained from contrast sensitivity formulations of retinal ganglion cells. The proposed quality estimator \textbf{CSV} is tested on the LIVE, the Multiply Distorted LIVE, and the TID 2013 databases, and it is always among the top two performing quality estimators in terms of at least ranking, monotonic behavior or linearity.
\end{abstract}

\end{frontmatter}

\section{Introduction}
\label{sec:intro}
Image quality estimators are designed to evaluate one or more dimension of quality of experience (QoE) of the end user. We usually model the QoE by considering the constraints caused by the physical systems including but not limited to the inherent digital nature of acquisition, compression,  streaming, and display. Storage and transfer of visual data can lead to noticeable degradation. Moreover, physical appearances and perceived appearances are different from each other because of the processing within a visual system. Therefore, we need to consider a broad spectrum of degradation operations to design a realistic and standalone quality estimator.

An intuitive method to measure the quality of an image is to directly compare it with a pristine image, if available. Mean squared error (MSE) and peak signal-to-noise ratio (PSNR) are commonly used in the literature to measure pixel-wise fidelity. These methods are preferred because of their simplicity. However, they ignore the perceived quality by focusing only on the fidelity. Characteristics based on human visual system (HVS) are used to extend PSNR to obtain more perceptual quality estimators, which are
denoted as PSNR-HVS \cite{Ponomarenko2007}, PSNR-HVS-M \cite{Ponomarenko2007}, PSNR-HA \cite{Ponomarenko2011}, and PSNR-HMA \cite{Ponomarenko2011}. 

In addition to pixel-wise SNR methods,  structural methods are also used to estimate the quality of images. The authors in \cite{Wang2004} propose a full-reference method (SSIM) quantifying the changes in luminance, contrast, and structure in the spatial domain. These structure-based methods are also extended to multi-scale (MS-SSIM) \cite{Wang2003}, complex domain (CW-SSIM) \cite{Inter2005}, and  information-weighted (IW-SSIM) \cite{Wang2011} versions. Multi-scale representations and transforms are used in the modeling of a visual system inspired by the finding that neural responses in a visual cortex perform scale-space orientation decomposition. In addition to structural similarity, feature similarity index (FSIM), which utilizes phase congruency (PC) and gradient magnitude (GM), is also used in the literature \cite{Zhang2011}. Alternatively, characteristics of source images can also be utilized to obtain a more accurate quality assessment. The authors in \cite{Mittal2012} propose a no-reference image quality assessment method based on natural image statistics in the spatial domain. In addition to modeling the source characteristics, the authors in \cite{Sheikh2006b} also model distortion and HVS characteristics.

The HVS is more sensitive to changes in intensity compared to color as exploited in the chroma subsampling for image coding \cite{Lambrecht2001}. Therefore, luma channels are more informative compared to chroma channels in terms of perceived quality. Although color may not be as informative as intensity, there is still additional information in color that is not conveyed by intensity.
An intuitive way to introduce color perception is pixel-wise fidelity as in PerSIM \cite{Temel2015} and FSIMc \cite{Zhang2011}. These pixel-wise fidelity approaches overlook the inherent structure of color, which implies that color is not a metric space  and when it is treated as such, it would lead to problems \cite{Kinsman1980}. The difference between individual color channels would not necessarily correspond to the perceived difference between colors. Therefore, instead of treating color channels as equivalent and separate, we should focus on the overall perceived color as a combination of these channels and calculate the color difference. In terms of the application field of the color difference equations, the approach  in \cite{Zhang2013} is a transition from basic tone matching to textured image comparisons. The authors in \cite{Johnson2003} discuss the connections between image quality, appearance, and color difference. In \cite{Yang2012}, the authors combine \texttt{CIEDE2000} color difference with the printing industry standards for visual verification to assess perceived image quality.

Color difference equations are commonly used in the literature for tone matching applications. Although they are used to assess the quality of images, the restrictions behind the design of color difference equations have been overlooked.  \texttt{CIEDE2000} is designed to quantify the differences between similar color tones. The authors in \cite{Werman2012} explain how basic color difference equations can fail in case of significant color changes and suggest quantifying the distance between color labels. In order to measure the distance between color labels, a transportation formula is used, which is known as Earth Mover's Distance (\texttt{EMD}). \texttt{EMD} calculates the minimal cost required to transfer one distribution into the other until they are equivalent \cite{Rubner2000}.  The authors in \cite{Werman2012} calculate the \texttt{EMD} between color naming labels  using the weighted flow based on the perceived color distance and the distance is used to perform color-based edge detection. The authors in \cite{Temel2014} use the combination of the \texttt{CIEDE2000} color difference and the transportation distance between color name labels to quantify the perceptual color difference for image quality assessment.

The authors in \cite{Norm_Book_2001} investigate the functional role of neurons and neural systems. More specifically, they try to develop a model for early sensory processing, which includes nonlinearities and adaptation mechanisms in cortical neurons. The statistics of natural scenes can be analyzed by decomposing images using basis functions. Intuitively, natural images cannot be decomposed into independent components using linear basis functions because the origin of these images are not based on fusion of independent patterns. Even individual patterns can be represented as combinations of linear basis functions, linearities turn into nonlinearities in case of occlusion. Therefore, linear decomposition-based representations can only approximate natural scenes. The statistical properties of natural scenes can be extracted using steerable pyramid \cite{Simoncelli1992} whose basis functions are translations, rotations, and dilations of a common filter kernel. When natural images are projected onto these basis functions, the joint statistics of these coefficients contain non-linear dependencies. It has been shown that these non-linear dependencies can be reduced using normalization operations \cite{Simoncelli1997,Simoncelli1999,Wainwright1999}. These normalization operations correspond to the suppression mechanisms in a visual system \cite{Simoncelli1999}. Divisive normalization transform is used in image quality assessment methods to reduce the effect of these redundancies   \cite{Li2009,Gol2015}. Alternatively, normalization can be directly performed in the spatial domain. In the learning-based computer vision applications, images are fed to normalization blocks to filter out redundancies and keep distinctive features. A commonly used architecture in these learning methods is convolutional neural network (CNN), which contains normalization layers. When CNNs are used for object recognition, a global normalization is applied over an entire image to avoid saturation, illumination, and contrast variation issues. In case of image quality assessment, local normalization outperforms global normalization \cite{Kang}.

In image processing and computer vision literature, difference of Gaussian and Laplacian of Gaussian  operators are commonly used to obtain image descriptors. These descriptors can extract the band-pass information that characterizes images in a more distinctive way compared to original pixel values. Therefore, they are commonly used in applications including but not limited to classification and image retrieval. The authors in \cite{Robson1966} show that the contrast sensitivity of retinal ganglion cells of a cat can be modeled with a difference of Gaussian formulation. Similarly, a Gaussian derivative-like approach is proposed by the authors in \cite{Young1987} to model the neural mechanism in a human foveal retinal vision and it is claimed to outperform Gabor filters based on  model-free Wiener filter analysis. The authors in \cite{Temel2015} use Laplacian of Gaussian operators to partially mimic the role of a visual system in perceptual image quality assessment.

In this work, we propose a full-reference image quality estimator that combines color difference, color name distance, structural difference, and retinal ganglion cell-based difference blocks.  An overview of the proposed quality estimator is given in Section \ref{sec:csv_overview}. Perceptual color difference is explained in Section \ref{sec:color}, which is based on color difference formulations and color name distances. Structural difference is explained in Section \ref{sec:structure} and retinal ganglion cell-based difference is described in Section \ref{sec:visual}. 
Spatial pooling, parameter tuning, and complexity analysis are discussed in Section \ref{sec:pooling_tuning}, and validation of the proposed quality estimator is given in Section \ref{sec:validation}. Finally, we conclude our work in Section \ref{sec:conclusion}.

\section{INTRODUCTION TO CSV}
\label{sec:csv_overview}
\begin{figure*}[ht!]
\begin{adjustwidth}{-2.5cm}{}
\centering
\includegraphics[width=1.1\linewidth]{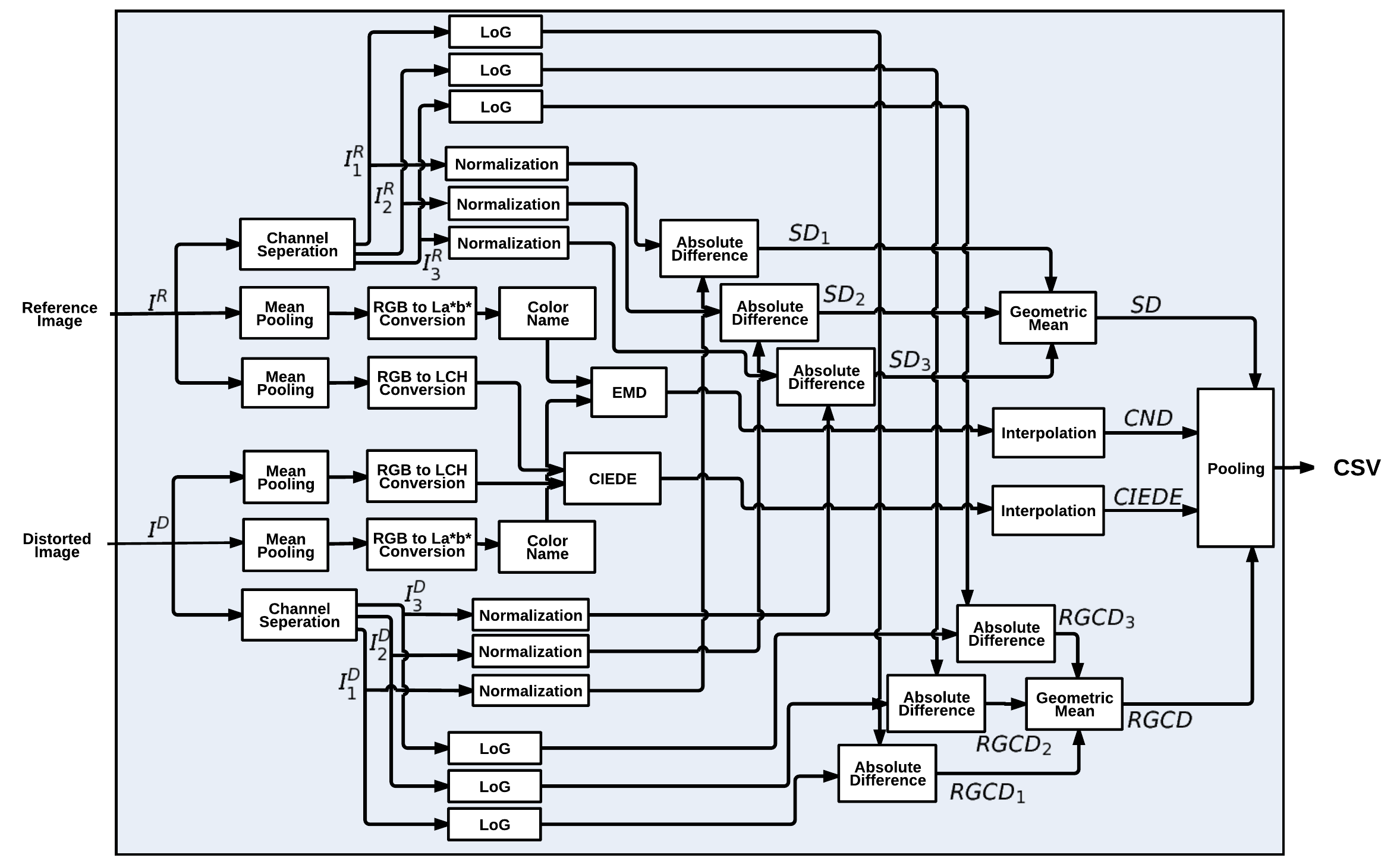}
\caption{CSV pipeline.}
\label{fig:CSV_Pipeline}
\end{adjustwidth}{}
\end{figure*}

The proposed method \texttt{CSV} is a full-reference image quality estimator, whose pipeline is given in Fig. \ref{fig:CSV_Pipeline}. Initially, a reference ($I^R$) and a distorted ($I^D$) image are in the RGB color domain. First, color channels of these RGB images are separated and fed into Laplacian of Gaussian (\texttt{LoG})  and  normalization blocks. For each channel, the outputs of the \texttt{LoG} blocks are fed into absolute difference blocks to obtain retinal ganglion cell-based difference (\texttt{RGCD}) maps. Geometric mean of these maps are computed to obtain a final \texttt{RGCD} map. Similarly, separated RGB channels are fed into the normalization and the absolute difference blocks to obtain structural difference (\texttt{SD}) maps, which are combined with a geometric pooling operation to obtain a final \texttt{SD} map.

To obtain color difference and color name distance maps, compared images are mean pooled and transformed into the LCH and the La*b* color domains, in which chroma and luma channels are separated. The mean pooled La*b* maps are fed into color name blocks to obtain color descriptors. The Earth Mover's Distance (\texttt{EMD}) between descriptors is calculated for each pixel to obtain a color name distance map (\texttt{CND}), which is interpolated to the same size with the compared images. Mean pooled LCH maps are fed into \texttt{CIEDE} blocks to obtain a color difference map. This difference map is interpolated to the same size with the compared images to obtain a \texttt{CIEDE} map. Finally, all feature maps are pooled together to obtain an estimated quality score, which is denoted as \texttt{CSV}.

\begin{figure*}[htbp!]
\begin{minipage}[b]{0.45\linewidth}
  \centering
\includegraphics[width=\linewidth, trim= 25mm 80mm 25mm 80mm]{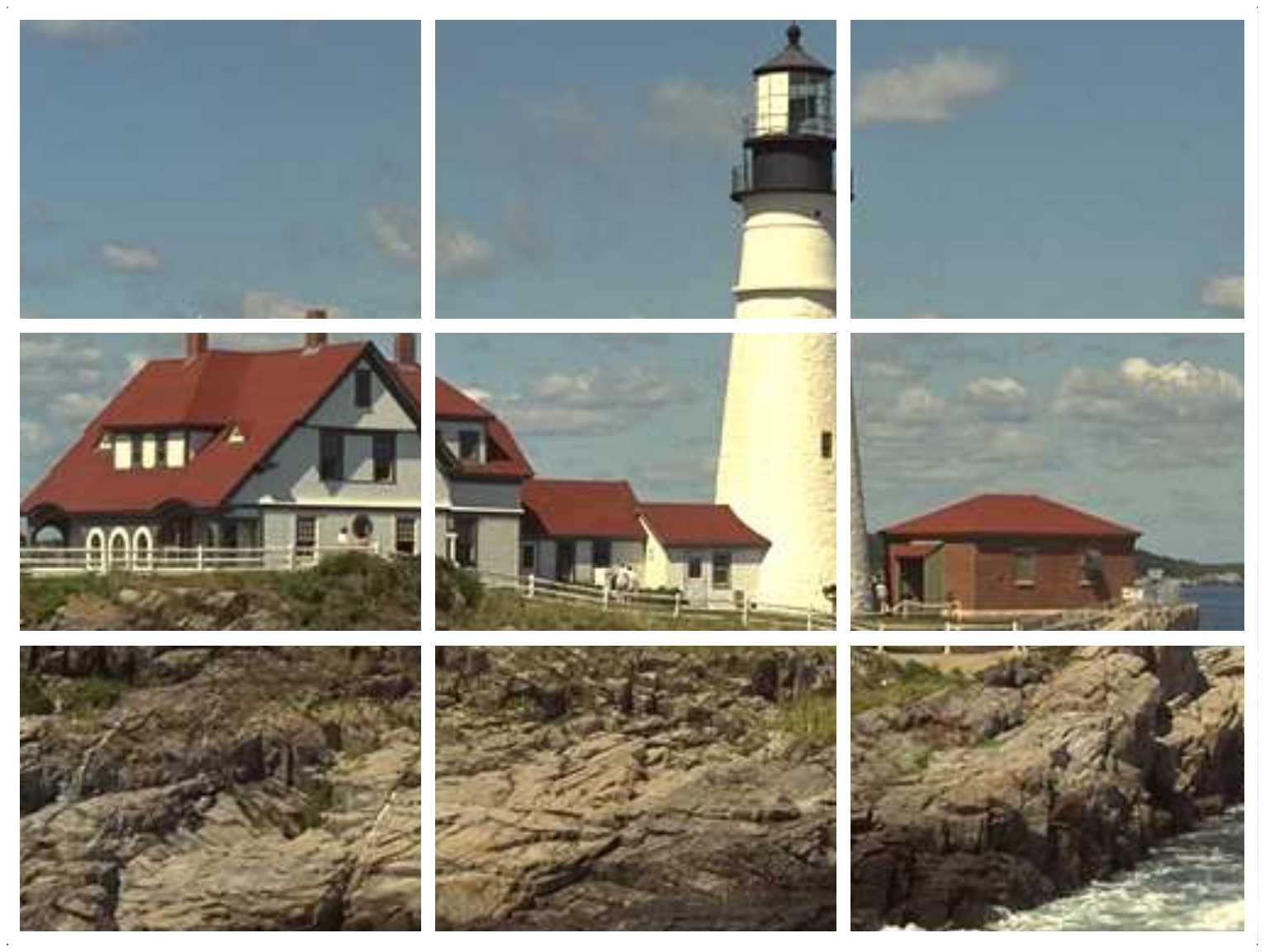}
  \vspace{0.03cm}
  \centerline{\footnotesize{(a)Reference Image}}
    \vspace{-0.35cm}
\end{minipage}
  \vspace{0.20cm}
\hfill
\begin{minipage}[b]{0.45\linewidth}
  \centering
\includegraphics[width=\linewidth, trim= 25mm 80mm 25mm 80mm]{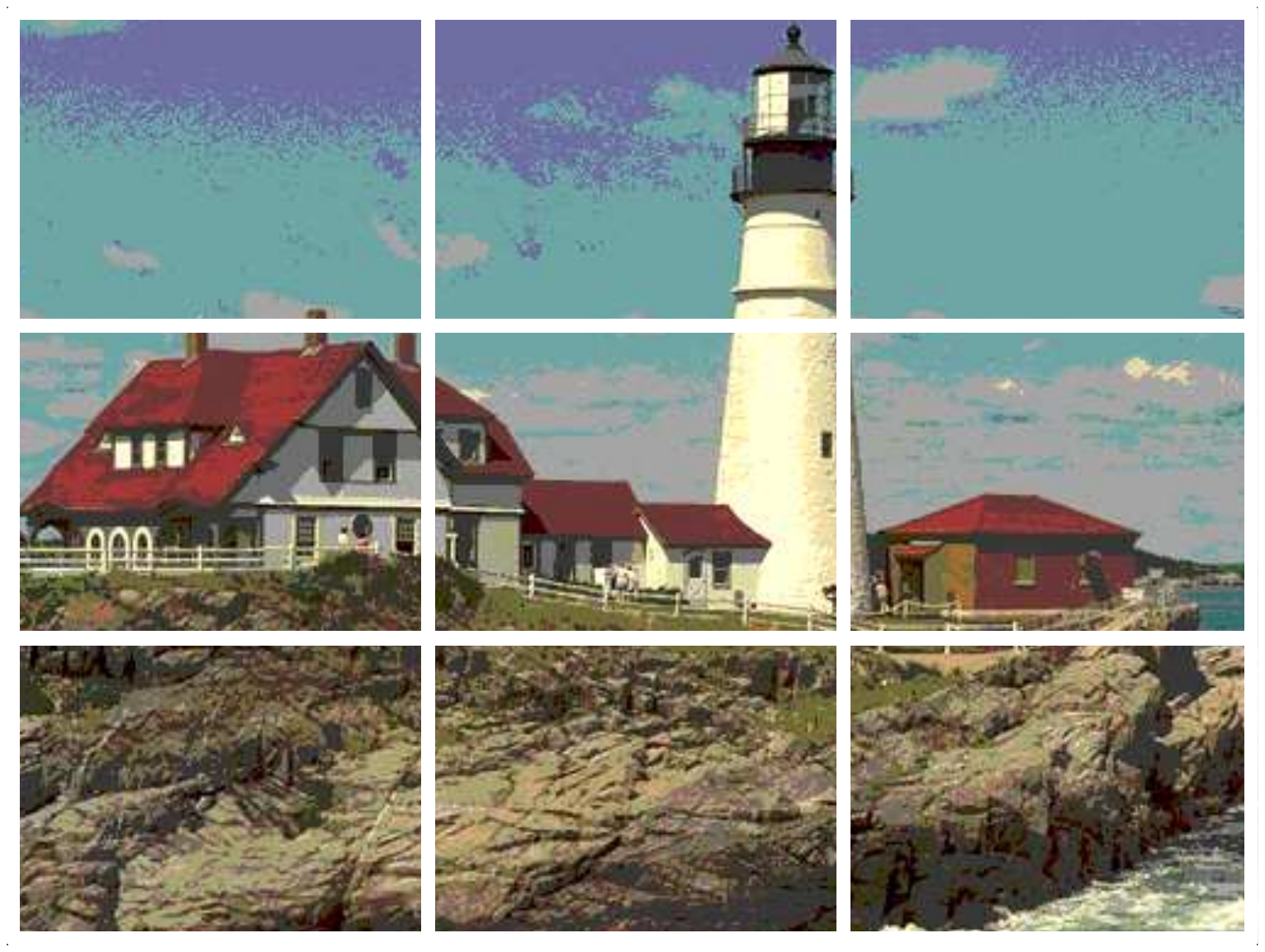}
  \vspace{0.03cm}
  \centerline{\footnotesize{(b) Distorted Image }}
    \vspace{-0.35cm}
\end{minipage}

\caption{Reference and distorted images. }\vspace{-.5cm}
\label{fig:reference_distorted}
\end{figure*}

In the following sections, we use the \texttt{lighthouse2} image, which is from the TID 2013 database \cite{tid13journal}, degraded with quantization   to illustrate the distortion maps corresponding to the output of each individual block in \texttt{CSV}. All of the images and the feature maps are shown with a grid structure to make the visual comparison easier. The reference and the distorted images are shown in Fig. \ref{fig:reference_distorted}. Structural degradations over the sky region are obvious around the top grids and also around the right side of the middle row. There is a significant color degradation around the upper part of the top rows and also observable tone difference between other sky regions. Degradations are less observable around the highly textured regions as observed in the bottom grids, where we have the textured rock components. In the middle row, we can observe degradation over the roof of the houses and around the windows, where we have edges or sharp transitions. However, it is not easy to observe degradations in regions with over exposure such as the surface of the lighthouse.

\section{Perceptual Color Difference (PCD)}
\label{sec:color}

\subsection{Color Difference (CIEDE)}
\label{subsec:color_ciede}
The \texttt{CIEDE2000} formulation combines lightness, chroma, and hue difference equations in the LCH color space. To compute CIEDE2000, we convert RGB images into LCH images and perform mean pooling over compared images separately using non-overlapping windows. Mean pooling operation is formulated as
\begin{equation}
\label{eq:mean}
\mu_{i,j}^R=\frac{1}{W^2}
\sum_{m=m_0+1}^{ m_0+W} \sum_{n=n_0+1}^{ n_0+W} I_j^R[m,n], 
\end{equation}
where $m_0$ and $n_0$ are the top left coordinate of each window, $m$ and $n$ are the pixels indices, $W$ is the window size, $j$ is the color channel index, $i$ is the window index that depends on $m_0$ and $n_0$, and $R$ indicates the type of the image as reference. In case of a distorted image, the $R$ superscript is replaced with the $D$ superscript. We calculate the CIEDE2000 difference \cite{Yang2012} pixel-wise between the reference ($\mu^R$) and the distorted ($\mu^D$) mean pooled maps. The CIEDE2000 formulation includes environmental tuning parameters $K_L$, $K_C$, and $K_H$, weighting factors $S_L$, $S_C$, and $S_H$ that are functions of l, a*, or b* color channel values, and a rotation factor that is a non-linear function of a* and b* color channels. A detailed description of the rotation and the weighting factors can be found in \cite{Yang2012}. Finally, the \texttt{CIEDE} map is resized to the resolution of the compared images using a bicubic interpolation operation.

\subsection{Color Name Distance (CND)}
\label{subsec:color_cnd}
Color names are pixel-wise descriptors. In these descriptors, each entity corresponds to the probability of an input pixel being perceived as one of the $N$ basic colors. In case of $N=11$, the color names in a dictionary are: black, blue, brown, grey, green, orange, pink, purple, red, white, and yellow \cite{Berlin1969}. Google image search was used to obtain labeled images, which also include wrong labels. Variants of the probabilistic latent semantic analysis model were used over noisy data to obtain a color name lookup table \cite{VandeWeijer2009}. We denote $L()$ as the color name lookup operator that receives a pixel value and returns an $N$ dimensional color name descriptor. Color name distance map is given by
\begin{equation}
\label{eq:cnd_1}
CND_{i}=EMD(L(\mu_{i}^R),L(\mu_{i}^D)),
\end{equation}
where $\mu^R$ is the reference map, $\mu^D$ is the distorted map, and \texttt{EMD} is the Earth Mover's Distance operator. The color name distance map is resized to the input image resolution using a bicubic interpolation.

\begin{figure}[htbp!]
\begin{center}
\noindent
  \includegraphics[width=0.80\linewidth]{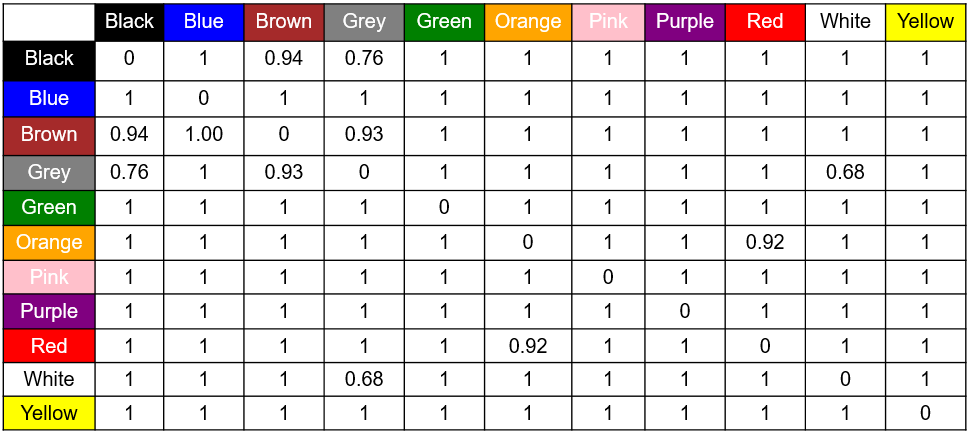}
    \vspace{-0.2cm}
  \caption{Perceptual color difference weights.}
  \label{fig:cnd_2}
\end{center}
\end{figure}

We use \texttt{EMD} to quantify the difference between color name descriptors  \cite{Werman2012} as
\begin{equation}
\label{eq:cnd_2}
EMD(L(\mu_{i}^R),L(\mu_{i}^D))=\underset{f_{k,l}}{min}\left \{ \sum_{k=1}^{N} \sum_{l=1}^{N}d_{i,k,l}f_{i,k,l} \right \}, 
\end{equation}
where $i$ is the window index, $k$ is the reference color name index, $l$ is the compared color name index, $N$ is the number of colors in the dictionary, $f_{i,k,l}$ is the flow from the $k^{th}$ color probability in the reference to the $l^{th}$ color probability in the compared descriptor for the $i^{th}$ window, and $d_{i,k,l}$ is the ground distance. Constraints of the flow equation \cite{Werman2012} are formulated as
\begin{equation}
\label{eq:cnd_3}
 \sum_{k=1}^{N} \sum_{l=1}^{N}f_{i,k,l}=1,\quad f_{i,k,l} \geq 0, 
\end{equation}
where sum of the overall flow adds up to unity and the flow is defined to be non-negative. We use perceptual color differences as the ground distance ($d$) in Eq. \ref{eq:cnd_2}. The joint distribution of basic color terms in the La*b* color space is used to obtain the perceptual differences between colors \cite{Werman2012}, which are summarized in  Fig. \ref{fig:cnd_2}. If we compare two entities that have the same color index, the perceptual difference is zero. When colors are perceptually similar such as black-brown, black-grey, brown-grey, grey-white, and orange-red, weights are in between zero and one.
 

%
%

\subsection{Color Difference Measurement under Significant Degradations} 
\label{subsec:color_ciede_cnd}
\begin{figure}[htbp!]
\begin{center}
\noindent
  \includegraphics[width=0.3\linewidth,]{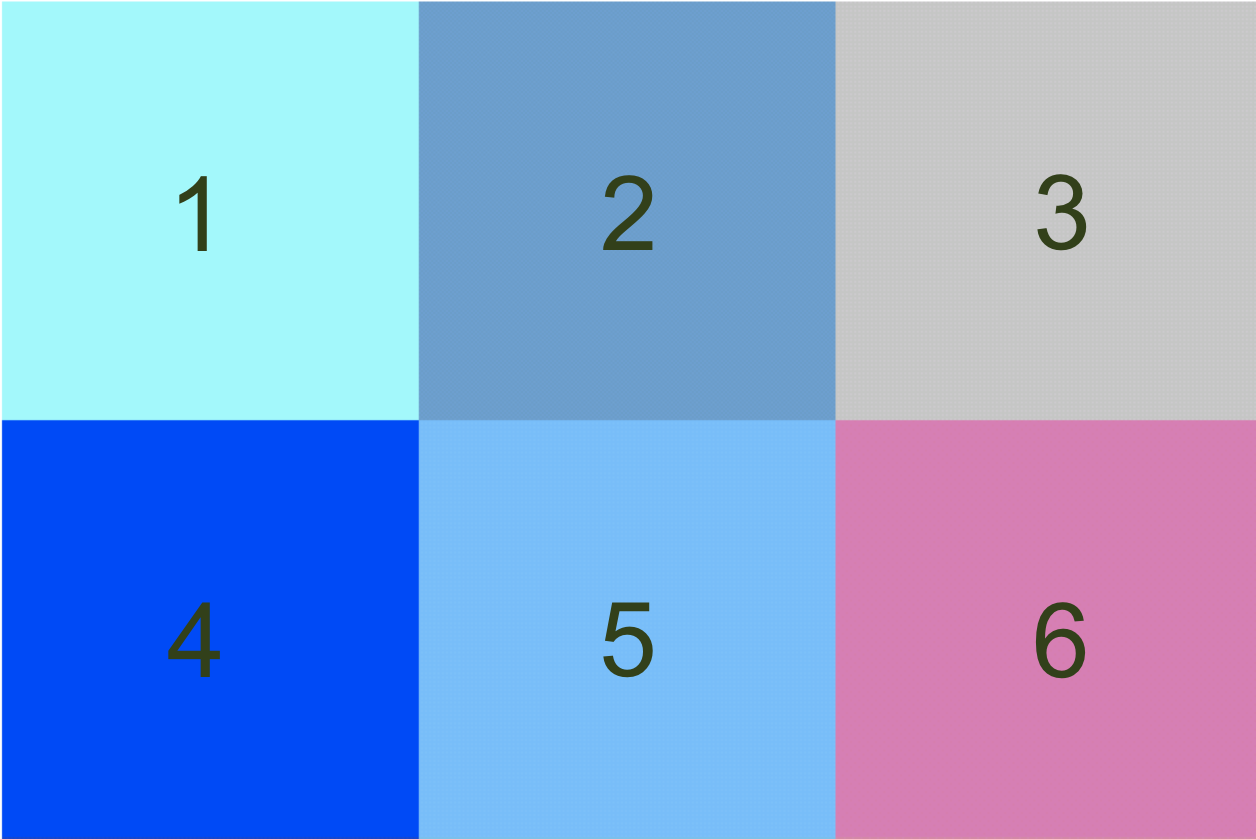}  
  \caption{Color difference chart.}
  \label{fig:pcd_1}
\end{center}

\end{figure}

To test the robustness of  \texttt{CIEDE} and \texttt{CND} under significant color degradations, we measure the difference between color tones given in Fig. \ref{fig:pcd_1}. There are six different color tones labeled from one to six. First column corresponds to reference colors, second column contains perceptually similar colors, and third column contains colors that are perceptually less similar. We compare the reference colors with the ones in the same row so there are two comparisons in each row and four in total. For each comparison, we provide the \texttt{CND} and the \texttt{CIEDE2000} values in Table \ref{tab:pcd_1}.

 \begin{table}[htbp]
  \centering
  \footnotesize
    \vspace{-10.0mm}
  \caption{The CND and the CIEDE2000 values between the reference and the compared colors.}  
    \begin{tabular}{|c||c|c||c|c|}
    \hline
 \multirow{2}[3]{*}{} &\multicolumn{2}{c||}{\textbf{CND}} & \multicolumn{2}{c|}{\textbf{CIEDE2000}} \\ \cline{2-5}
    \texttt{} & $2$ & $3$ & $2$ & $3$ \\ \hline
    \texttt{$1$}  &0.38  &0.70  &27.96  &27.27  \\ \hline
 \multirow{2}[3]{*}{} &\multicolumn{2}{c||}{\textbf{CND}} & \multicolumn{2}{c|}{\textbf{CIEDE2000}} \\ \cline{2-5}
    \texttt{} & $5$ & $6$ & $5$ & $6$ \\ \hline   
    \texttt{$4$}  &0.25  &0.97  &51.55  &47.42  \\ \hline
    \end{tabular}%
    \vspace{-2.0mm}

  \label{tab:pcd_1}
\end{table}

The \texttt{CND} values between the reference and the similar tones significantly differ from the \texttt{CND} values  between the reference and the different tones. However, it is not easy to differentiate between the \texttt{CIEDE2000} values. Moreover, perceptually different tones lead to lower \texttt{CIEDE2000} values. \texttt{CIEDE2000} is designed for small scale color differences and it potentially overlooks the differences between colors under significant degradations. Therefore, we can use \texttt{CND} to complement \texttt{CIEDE2000}.

\subsection{Visualization of CIEDE and CND Maps}
\label{subsec:color_ciede_cnd_vis}
\begin{figure*}[htbp!]

\begin{minipage}[b]{0.45\linewidth}
  \centering
\includegraphics[width=\linewidth, trim= 25mm 80mm 25mm 80mm]{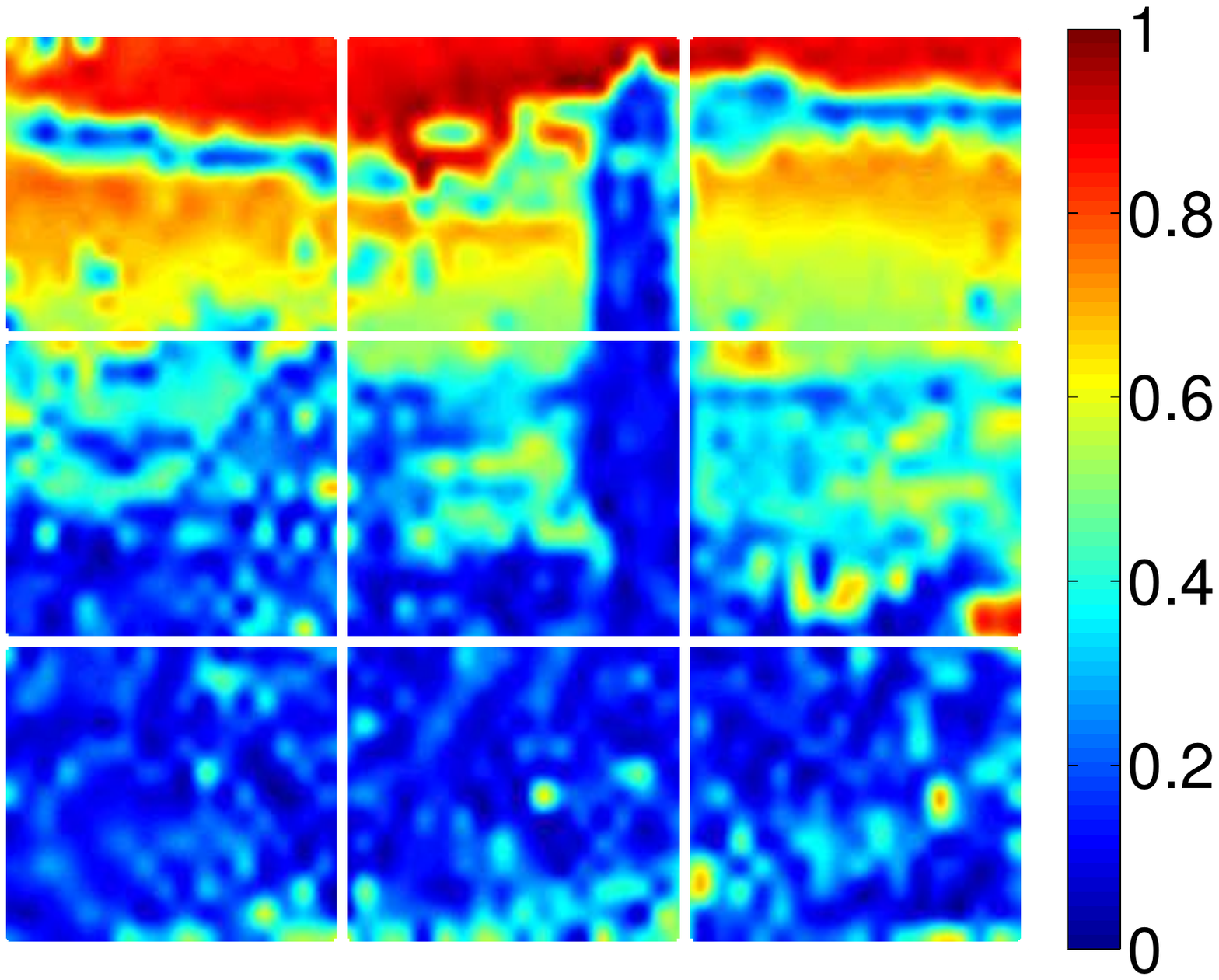}
  \vspace{0.03 cm}
  \centerline{\footnotesize{(a) CIEDE2000 (\texttt{CIEDE}) map  } }
    \vspace{-0.35cm}
\end{minipage}
  \vspace{0.20cm}
\hfill
\begin{minipage}[b]{0.45\linewidth}
  \centering
\includegraphics[width=\linewidth, trim= 25mm 80mm 25mm 80mm]{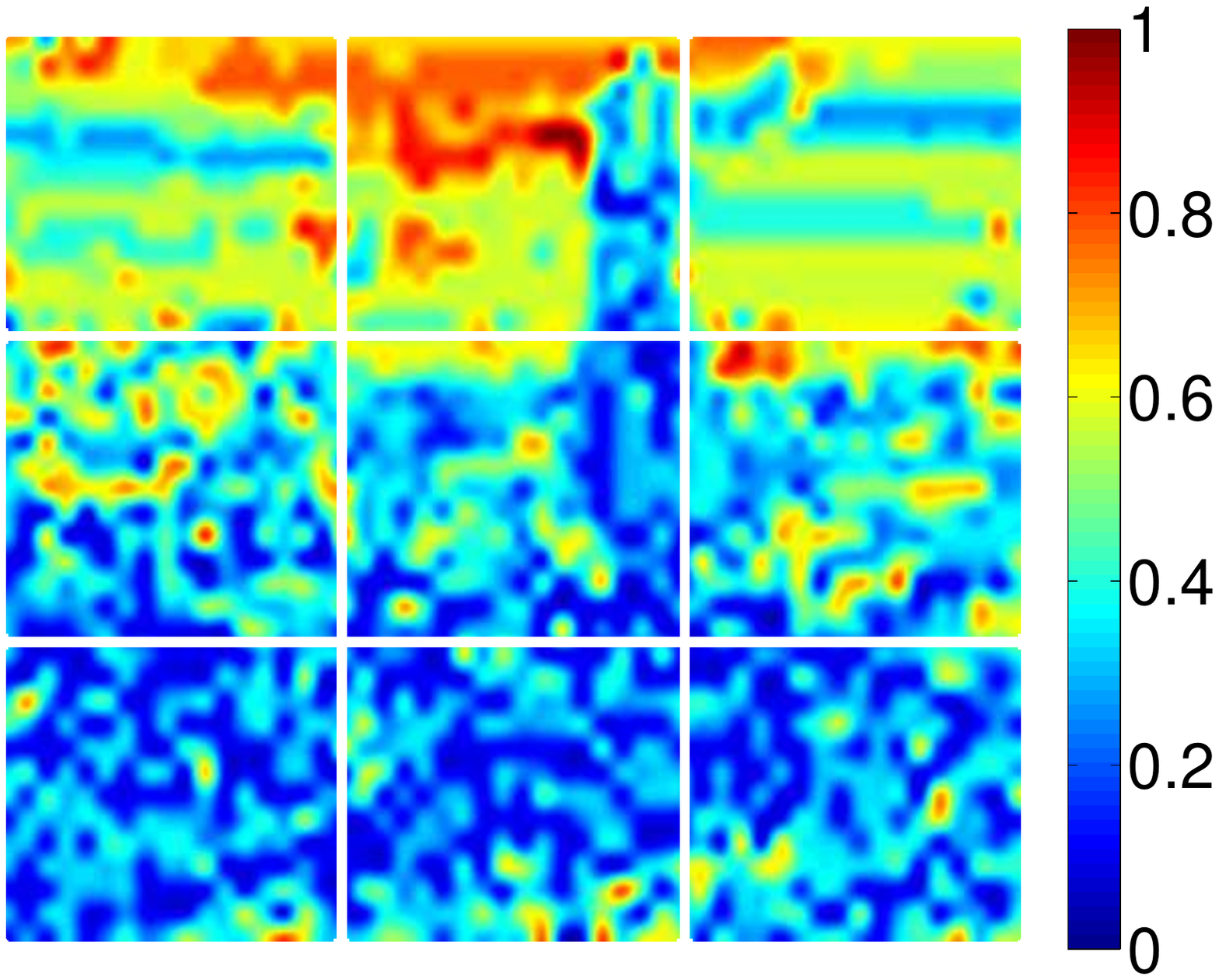}
  \vspace{0.03 cm}
  \centerline{\footnotesize{(b) Color name distance (\texttt{CND}) map }}
    \vspace{-0.35cm}
\end{minipage}
  \vspace{0.20cm}

\caption{Color-based difference and distance maps.}\vspace{-.5cm}
\label{fig:visual_color}
\vspace{-1.0mm}
\end{figure*}

\texttt{CIEDE} and \texttt{CND} maps corresponding to the images in Fig. \ref{fig:reference_distorted} are shown in Fig. \ref{fig:visual_color}.  These maps are distortion maps that lead to high values for significant degradations and vice versa. Color goes from blue to red under significant degradation as shown in the color bar.  Both of the color-based methods detect the degradations around the sky and the clouds as the most disturbing part of the image, which are shown in Fig. \ref{fig:visual_color}. Distortion levels around the textured regions are detected as either low or mediocre. \texttt{CIEDE} leads to a more consistent distortion map whereas \texttt{CND} is more sensitive to minor changes.

\section{Structural Difference (SD)}
\label{sec:structure}
We measure the structural difference (\texttt{SD}) as the absolute difference between locally normalized feature maps. Mean values of non-overlapping windows are calculated as in Eq. \ref{eq:mean} and standard deviations of windows are given by
\begin{equation}
\label{eq:BSD2}
\sigma_{i,j}^R= \sqrt{\frac{1}{W^2} \sum_{m=m_0+1}^{m_0+W} \sum_{n=n_0+1}^{n_0+W} (I_j^R[m,n]- \mu_{i,j}^R)^2}, 
\end{equation}
where $\mu^R$ is the mean map, $\sigma^R$ is the standard deviation map, $W$ is the window size, $m_0$ and $n_0$ are the index of the top left coordinate of each window, $j$ is the color channel index, $i$ is the index of each window that depends on $m_0$ and $n_0$, the reference image is represented with $R$, and the distorted image with $D$.

\begin{figure}[htbp!]
\label{fig:SD}
  \centering
\includegraphics[width=0.45\linewidth, trim= 25mm 80mm 25mm 80mm]{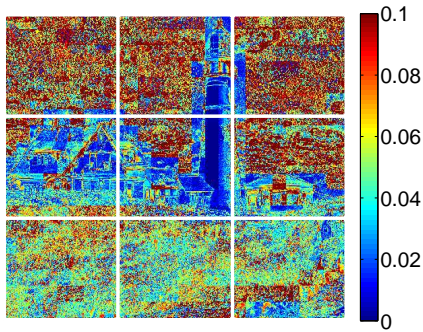}
\caption{Structural difference (\texttt{SD}) map.}\vspace{-.2cm}
\label{fig:visual_SD}
\vspace{-1.0mm}
\end{figure}

The normalization operation is composed of two steps. First, the local mean ($\mu_{i,j}^R$) is subtracted from each pixel in every color channel. Then, the mean shifted values are divided by the local standard deviation ($\sigma_{i,j}^R$) over each color channel for every pixel. Structural difference map is obtained  as the absolute difference between the normalized color channels ($\left| Norm(I_j^R)-Norm(I_j^D\right)|$). We combine the pixels of difference maps from each color channel to obtain a single quality map as given in Fig. \ref{fig:visual_SD}.

The block-wise nature of \texttt{SD} leads to discontinuities among non-overlapping windows, especially around the window borders as seen in Fig. \ref{fig:visual_SD}. Within each window, there are fluctuations and inconsistencies because of the pixel-wise nature. Sky and cloud regions lead to high \texttt{SD} values whereas textured regions such as rocks and buildings have lower \texttt{SD} values. Over-exposed regions around the houses and the lighthouse such as the walls lead to lower \texttt{SD} values compared to the regions with sharp transitions and edges such as the regions around the windows or the edges of the roofs.

\section{Retinal Ganglion Cell-based Difference (RGCD)}
\label{sec:visual}

 We can use a difference of Gaussian operator to formulate contrast sensitivity mechanisms of retinal ganglion cells in a visual system. Moreover, we can combine multiple difference of Gaussian operators to decompose a visual stimuli into various frequency bands. However, fusing the output of these operators would require parameter tuning. To avoid the tuning issue, we use the second derivative of a Gaussian operator, which is used to approximate the difference of Gaussian operator. Second derivative of a Gaussian operator corresponds to the Laplacian of Gaussian (\texttt{LoG}) operator, which is formulated as 
\begin{equation}
\label{eq:log_1}
LoG[m,n]= \frac{1}{\sqrt{2\pi\sigma ^2}}   \frac{m^2+n^2-2\sigma^2}{\sigma^4}  e^{-\frac{m^2+n^2}{2\sigma^2}}, 
\end{equation}
where $m$ and $n$ are the filter coordinates with respect to the center, and $\sigma$ is the standard deviation. Reference and distorted images are convolved with the \texttt{LoG} operator and the absolute difference is taken as
\begin{equation}
\label{eq:log_2}
RGCD_{i,j}=\left| I_{i,j}^R\asterisk LoG - I_{i,j}^D\asterisk LoG\right|,
\end{equation}
where $\asterisk$ is the convolution operator, $RGCD$ is the retinal ganglion cell-based difference map, $j$ is the color channel index, and $i$ is the window index.

\begin{figure}[htbp!]

  \centering
\includegraphics[width=0.45\linewidth, trim= 25mm 80mm 25mm 80mm]{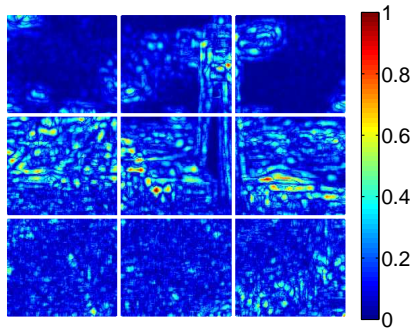}
\caption{Retinal ganglion cell-based difference (\texttt{RGCD}) map.}
\label{fig:visual_log}
\vspace{-1.0mm}
\end{figure}

The \texttt{LoG} block filters out the smooth and lowly textured regions and focuses on the perceived degradations around sharp transitions so \texttt{LoG} detects the degradation around houses, lighthouse, and clouds as shown in Fig. \ref{fig:visual_log}. It also detects some of the structural degradations around the sky and the rocks. \texttt{LoG} is not very sensitive to the level of color-based degradations as observed in the upper row sky region.

\section{Spatial Pooling, Parameter Tuning, and Complexity Analysis}
\label{sec:pooling_tuning}
 
\subsection{Spatial Pooling}
\label{subsec:pooling}
 The \texttt{RGCD} and the \texttt{SD} blocks are calculated over each color channel to detect visual degradations. Since numerical ranges of these quality estimators depend on the color distribution, arithmetic average over the color channels would be a biased estimator towards a highly populated color channel. Therefore, we calculate the geometric mean over the color channels to obtain final \texttt{RGCD} and \texttt{SD} values as
   \begin{equation}\label{eq:pool_sd}
SD_{i}=\sqrt[3]{SD_{i,1}\cdot SD_{i,2} \cdot SD_{i,3}},	
    \end{equation}
   \begin{equation}\label{eq:pool_log}
RGCD_i=\sqrt[3]{RGCD_{i,1}\cdot RGCD_{i,2} \cdot RGCD_{i,3}}.		
    \end{equation}

 Color-based (\texttt{CIEDE}, \texttt{CND}), structure-based (\texttt{SD}), and visual system-based (\texttt{RGCD}) maps need to be pooled to obtain a final quality estimate. If quality maps were normalized to the same range, we would be able to use additive fusion to obtain an estimate. However, normalizing quality maps based on the overall statistics of tested databases would not be a fair approach because the evaluation should not require any information other than the compared images. Therefore, we perform multiplicative fusion to combine the feature maps of the individual blocks as
   \begin{equation}\label{eq:csv_pool_1}
CSV_{i}=RGCD_{i} \cdot SD_{i} \cdot (A \cdot CND_{i}+(1-A) \cdot CIEDE_{i}),
    \end{equation}
where $A$ is the weight that scales the ranges of the \texttt{CIEDE} and the \texttt{CND} maps. All the quality estimator blocks are based on difference or distance operators that lead to high values in case of significant degradations. However, we want to propose a quality estimator that leads to high values in case of high quality and vice versa. Thus, we calculate the residual of the average distortion. To adjust the scalar range of \texttt{CSV}, the residual of the $P^{th}$ root is calculated to obtain the final quality score as
   \begin{equation}\label{eq:csv_pool_2}
CSV=1-\sqrt[P]{\frac{1}{M}\sum_{i=1}^{M} CSV_i}, 
    \end{equation}
where $i$ is the window index and $M$ is the number of windows in a reference or a distorted image.
 
\subsection{Parameter Tuning}
\label{subsec:tuning} 
The parameters of \texttt{CSV} are summarized in Table \ref{tab:param}. There are nine parameters, five of which ($K_L$, $K_C$, $K_H$, $T$, and $N$) are directly obtained from the original implementations of the used formulations. Two of the parameters ($W$ and $\sigma$) are selected from the visual assessment of quality maps. The parameter $P$ is selected based on the distribution of the quality estimates and the selection of the parameter $A$ is based on the color chart toy example.

 The \texttt{CIEDE} block has five parameters. $W$ is the size of the rectangular window set to $20\times20$ . $K_L$, $K_C$, and $K_H$ are the environmental parameters in the CIEDE2000 color difference equation set to $1.0$ in the CIE standard observation conditions. The subjective test setup cannot exactly match the standard conditions and these parameters need to be tuned based on the environment. Meanwhile, the proposed method should work for any image, independent of acquisition or display technology. Therefore, we fix these environmental parameters to standard values. We use a threshold ($T$) to limit the CIEDE2000 estimates to small degradations and set $T$ to $20$ as in \cite{Pele2009}. The window sizes ($W$) in \texttt{CIEDE},  \texttt{CND}, \texttt{SD}, and \texttt{RGCD} are set to $20\times20$ by visually assessing the distinctiveness of randomly selected feature maps. Smaller window size leads to the amplification of noise and fluctuations whereas larger window results in the loss of local information, both of which decrease the prediction accuracy. The standard deviation ($\sigma$) is set to $50$ in the retinal ganglion-cell based difference by visually assessing the feature maps.

\begin{table}[]
\centering
\tiny
\caption{The parameters in the proposed method \texttt{CSV}.}
\label{tab:param}
\begin{tabular}{|l|l|l|}
\hline
\textbf{CSV Block}                & Parameter & Value     \\ \hline
\multirow{5}{*}{\textbf{CIEDE}}   & $W$       & $[20,20]$ \\ \cline{2-3} 
                                  & $K_L$     & $1.0$     \\ \cline{2-3} 
                                  & $K_C$     & $1.0$     \\ \cline{2-3}
                                  & $K_H$     & $1.0$     \\ \cline{2-3}
                                   
                                  & $T$     & $20$     \\ \hline
\multirow{2}{*}{\textbf{CND}}     & $W$       & $[20,20]$ \\ \cline{2-3} 
                                  & $N$       & $11$      \\ \hline
\textbf{SD}                       & $W$       & $[20,20]$ \\ \hline
\multirow{2}{*}{\textbf{RGCD}}    & $W$       & $[20,20]$ \\ \cline{2-3} 
                                  & $\sigma$  & $50$      \\ \hline
\multirow{2}{*}{\textbf{Pooling}} & $A$       & $0.9$     \\ \cline{2-3} 
                                  & $P$       & $4$       \\ \hline
\end{tabular}
\end{table}

The researchers in \cite{Berlin1969} investigate the basic color terms in verbal usage and $20$ different languages are used to obtain universal categories independent of the language characteristics. As a consequence of these studies, the researchers defined $11$ basic color groups, which is used as the value of the parameter $N$. The difference between the color name descriptors are also computed with variations of L-norms and information theoretic formulations. However, the performance of the proposed quality estimator does not change significantly compared to Earth Mover's Distance. \texttt{CND} and \texttt{CIEDE} have different numerical ranges as shown in Table \ref{tab:pcd_1}. Therefore, we use weights to scale the color-based differences in the summation. The weight of  \texttt{CND} is set to $A$ and \texttt{CIEDE} to $(1-A)$, where $A$ is $0.9$. We select the remaining parameter A by finding the weight in the color chart example that assigns a higher value to CND compared to CIEDE and results in higher differences for less similar colors. If we assign a lower weight to CND ($0.8$ instead of $0.9$), weighted color difference formulation would not be able to detect the similar color tones in the second row.

 The quality score is obtained using Eq. \ref{eq:csv_pool_2} by computing the $P^{th}$ root of the average \texttt{CSV} value, where $P$ is set to $4$. The numerical range of \texttt{CSV} can be set to different values by using other monotonic functions or power values. However, scaling does not bias the performance of the quality estimation  because ranking-based performance metrics are not affected from the monotonic mapping, and regression would eliminate the effect of mapping in terms of linearity. Therefore, parameter selection process is independent from the performance validation stage and this independence should eliminate overfitting to the tested databases.

\subsection{Complexity Analysis}
\label{subsec:complexity}
We classify the main blocks in the proposed method according to their computational complexity. Channel separation, mean pooling, local normalization, absolute difference, geometric mean, and pooling operations are less computationally demanding compared to color space transformation, color name extraction, interpolation, EMD, CIEDE2000, and LoG filtering. We perform mean pooling over compared images with a $20\times20$ non-overlapping window to decrease the number of processed pixels by $400$ times, and we perform most of the computationally intensive operations after size reduction. Moreover, we use a robust version of EMD \cite{Pele2009} that is shown to be faster than the original version \cite{Rubner2000} up to $75-700$ times in various applications. We estimated the quality scores using the three best performing methods in the LIVE, the MULTI, and the TID databases, which contain more than $4,000$ images. We used a PC with a 3.50 GHz Intel(R) Core(TM) i7-3770K CPU, a 32 GB RAM, and a 64-bit operating system. When we compare the quality estimators in terms of the average time required to estimate a quality score, \texttt{CSV} is slower than FSIMc, and IW-SSIM. On average, the processing time per image is $0.18$ second for FSIMc, $0.35$ second for IW-SSIM, and $0.70$ second for \texttt{CSV}. 

We can further reduce the computational time of \texttt{CSV} by modifying the interpolation method, the filtering operation, and the data processing mechanisms. The interpolation method does not significantly affect the performance of the proposed method, as discussed in Section \ref{sec:validation}. Therefore, a bilinear- or a nearest-neighbor-based interpolation can be used instead of a bicubic interpolation to reduce the overall computational complexity. The Laplacian of Gaussian operator can be approximated with a difference of Gaussian operator, which can reduce the computational complexity. EMD and CIEDE values are computed for each pixel sequentially, which lead to $1,014$ processes in the LIVE, $520$ processes in the TID, and $2,304$ processes in the MULTI databases per image. These sequential processes can be parallelized to reduce the computation time.

\section{Validation}
\label{sec:validation}

The proposed method \texttt{CSV} is validated in commonly used databases LIVE \cite{Bovik2006}, Multiply Distorted LIVE (MULTI) \cite{Jayaraman2012}, and TID2013 (TID) \cite{tid13journal}. The LIVE database consists of $29$ reference images that are degraded with JPEG, JPEG2000 (Jp2k), White Noise  (Wn), Gaussian blur (Gblur), and simulated Fast-Fading Rayleigh channel errors (FF) to obtain $779$ distorted images \cite{Bovik2006}. In the MULTI database, there are two main distortion groups. The first group contains images that are firstly blurred and then compressed with JPEG whereas in the second group, images are blurred and then degraded with additive noise that has a standard normal pdf in each color channel. There are 405 multiply distorted images with 15 reference images \cite{Jayaraman2012}. The TID database contains $25$ reference images that are obtained from Kodak Lossless True Color Image Suite. All of these images are degraded with $24$ different distortion types and there are $5$ different levels for each distortion type. Thus, there are $3000$ ($25\times24\times5$) distorted  images in the TID database \cite{tid13journal}. Distorted images are grouped into six categories as noise, actual, simple, exotic, new, and color. Some of the categories contain the same distortion types. However, these distortion types are only counted once when the results are provided for the full database \cite{tid13journal}.

Objective image quality estimators assign a score to images according to the numerical ranges they are defined. Since the range of the objective and the subjective scores are not necessarily same, regression methods are used for a fair comparison of the quality estimators as introduced in \cite{Bovik2006}. The regression equation is given by
\begin{equation}
\label{eq:nonlinreg}
S=\beta_1 \left ( \frac{1}{1}-\frac{1}{2+exp(\beta_2(S_0 -\beta_3 ))} \right )+\beta_4 S_0 +\beta_5,
\end{equation}
where $\beta_1$ to $\beta_5$ are the parameters tuned to maximize the correlation between the subjective scores and the objective quality estimates, $S_0$ is the original quality estimate, and $S$ is the new quality estimate after regression. We follow the same approach with the authors in \cite{Bovik2006} and calculate the Pearson linear correlation coefficient (PLCC) after the regression operation for the LIVE and the MULTI databases. Moreover, we also provide ranking-based Spearman (SCC) and Kendall (KCC) correlation coefficient results. In the TID database, the results are provided in terms of monotonic behavior (SCC, KCC) as given in the online benchmark \cite{tidDATABASE}.

\begin{center}

\begin{table}[htbp!]

  \centering
    \tiny
    \caption{Performance of the methods in the LIVE database.}

    \begin{tabular}{|c||c|c|c|c|c|c|}   \hline

    \multirow{2}[3]{*}{\textbf{Sequence}}

          & \textbf{Jp2k}& \textbf{JPEG}& \textbf{Wn}& \textbf{Gblur}& \textbf{FF}  & \textbf{All}   
          
 \\ \cline{2-7}

    & \multicolumn{6}{c|}{\textbf{Pearson (PLCC)}}               \\ \hline     
    
	\textbf{CSV} &\bf 0.983 &\bf 0.978 &\bf 0.986  &\textbf{0.971}  &0.949 &\textbf{0.967}  \\ 
	\textbf{$CSV_b$} &\bf 0.000&\bf 0.000&\bf 0.000& -0.007&-0.004&\bf -0.001  \\ 
	\textbf{$CSV_n$} &\bf 0.000&\bf 0.000&\bf 0.000& -0.007&-0.004&\bf -0.001  \\  
	\textbf{$CSV_e$} &-0.008&-0.014&-0.001&\bf 0.000&-0.002&-0.034  \\ \hline
	\textbf{CIEDE}  &-0.035&-0.025&-0.004&-0.028&-0.085&-0.051\\ 

	\textbf{CND}  &-0.028&-0.012&-0.020&-0.043&-0.084&-0.044\\ 

	\textbf{RGCD}  &-0.005 &-0.004 &\textbf{+0.002}  &-0.014  &-0.023 &-0.005 \\
	\textbf{SD}  &-0.021&-0.023&-0.030&\ -0.008&\textbf{+0.031}&-0.021   \\ 
	\textbf{RGCD$\cdot$SD}  &-0.002&-0.006&-0.009&\textbf{+0.006}&\bf +0.006&-0.003   \\ \hline

	\textbf{MSE} &-0.048&-0.043&-0.020&-0.098&-0.049&-0.055\\

  \textbf{SSIM} &-0.020&-0.020&-0.010&-0.031&\textbf{+0.007}&-0.022\\

  \textbf{MS-SSIM} &-0.020&-0.017&-0.008&-0.028&-0.001&-0.020\\

 \textbf{IW-SSIM} &-0.023&-0.019&-0.005&-0.014&+0.003&-0.015\\ 

 \textbf{CW-SSIM} &-0.056&-0.050&-0.037&-0.203&-0.114&-0.095\\

 \textbf{FSIMc} &-0.023&-0.029&-0.009& -0.015&+0.003&-0.018\\ 

 \textbf{PSNR-HA} 
&-0.001&-0.008&\textbf{0.000}&-0.023&+0.003&-0.008\\ \hline

    & \multicolumn{6}{c|}{\textbf{Spearman}}           \\ \hline    
 
 	\textbf{CSV} &\textbf{0.980}&0.960&\textbf{0.991}&0.973&0.937&0.959 \\ 
	\textbf{$CSV_b$} &\bf 0.000&0.000&\bf 0.000&-0.006&-0.006&-0.001  \\ 
	\textbf{$CSV_n$} &\bf 0.000&0.000&\bf 0.000&-0.006&-0.005&-0.001 \\  
	\textbf{$CSV_e$} &-0.003&-0.004&\bf +0.002&-0.008&+0.011&-0.011  \\  \hline
	\textbf{CIEDE}  &-0.011 &-0.002 &-0.007  &-0.034  &-0.117 &-0.060   \\ 
	
	\textbf{CND}  &-0.021 &-0.005 &-0.015 &-0.084  &-0.122  &-0.064     \\	

	\textbf{RGCD}  &-0.010 &-0.007 &-0.001  &-0.025  &-0.027 &-0.009   \\

	\textbf{SD}  &-0.026&-0.022&-0.024&-0.017&\textbf{+0.034}&-0.030 \\

	\textbf{RGCD$\cdot$SD}  &\textbf{+0.001}&0.000&\textbf{+0.002}&\textbf{+0.004}&+0.009&\textbf{+0.001}   \\ \hline

 	\textbf{MSE} &-0.026 &-0.028 &\textbf{0.000} &-0.100 &-0.001 &-0.049 \\  
	
	  \textbf{SSIM} &0.000 &\textbf{+0.001} &-0.009 &-0.001 &\textbf{+0.036} &-0.009 \\ 
  \textbf{MS-SSIM} &0.000 &\textbf{+0.002} &-0.007 &0.000 &+0.031 &-0.007 \\ 

 \textbf{IW-SSIM} &-0.001 &0.000 &-0.010 &\textbf{+0.010} &+0.029 &\textbf{+0.001}  \\ 

 \textbf{CW-SSIM} &-0.034 &-0.017 &-0.008 &-0.117 &-0.052 &-0.056 \\

 \textbf{FSIMc} &\textbf{+0.001}&\textbf{+0.002}&-0.012&\textbf{+0.009}&\textbf{+0.033}&\textbf{+0.002}  \\ 

 \textbf{PSNRHA}  &-0.003&-0.001&+0.001&-0.011&+0.010&-0.013  \\ \hline

    & \multicolumn{6}{c|}{\textbf{Kendall}}       \\ \hline

    \textbf{CSV} &\textbf{0.885} &0.840 &0.927 &0.868 &0.806 &0.834 
      \\ 
      
	\textbf{$CSV_b$} & \bf 0.000 &+0.002&0.000&-0.015&-0.010&-0.002  \\ 
	\textbf{$CSV_n$} & \bf 0.000 &+0.002&0.000&-0.014&-0.010&-0.002 \\  
	\textbf{$CSV_e$} &-0.008&-0.014& \bf +0.010&-0.022&+0.016&-0.022  \\  \hline    

	\textbf{CIEDE}  &-0.026 &-0.005 &-0.029 &-0.066  &-0.147  &-0.100  \\ 
	
	\textbf{CND}   &-0.050 &-0.013 &-0.058 &-0.132 &-0.154  &-0.106    \\

	\textbf{RGCD}   &-0.025 &-0.024 &-0.005 &-0.059  &-0.040  &-0.018  \\ 

	\textbf{SD}   &-0.065&-0.048&-0.076&-0.034&+0.056&-0.058  \\

	\textbf{RGCD$\cdot$SD}  &\textbf{+0.004}&0.000&\textbf{+0.010}&\textbf{+0.010}&+0.016&\textbf{+0.003}   \\ \hline
 
 	\textbf{MSE} &-0.060 &-0.059 &-0.002 &-0.164 &-0.015 &-0.085\\ 	
	  \textbf{SSIM} &0.000 &+0.003 &-0.032 &-0.010 &\textbf{+0.065} &-0.019\\  
  \textbf{MS-SSIM}  &-0.001 &\textbf{+0.008} &-0.023 &-0.004 &\textbf{+0.057} &-0.015\\ 
 \textbf{IW-SSIM} &-0.003 &\textbf{+0.006} &-0.034 &\textbf{+0.027} &+0.056 &\textbf{+0.004} \\

 \textbf{CW-SSIM} &-0.084 &-0.040 &-0.032 &-0.178 &-0.089 &-0.102 \\

 \textbf{FSIMc} &\textbf{+0.006}&\textbf{+0.012}&-0.041 &\textbf{+0.026}&\textbf{+0.063}&\textbf{+0.004}  \\ 

 \textbf{PSNRHA} &-0.008&-0.008&\bf +0.009&-0.028&+0.005&-0.029  \\ \hline

    \end{tabular}%
  \label{tab:LIVE_Results}
\vspace{-6.0mm}
\end{table} 

\end{center}
     
\vspace{-12.0mm}

The results of the LIVE, the MULTI, and the TID databases are given in Table \ref{tab:LIVE_Results}, \ref{tab:Multi_Results}, and \ref{tab:TID_Results}. Three highest performing methods are highlighted in each category to illustrate the best performing quality estimator. In case of an equality, all the methods that perform similarly are highlighted. There are $14$ methods compared in the TID literature, whose results are published online \cite{tidDATABASE}. In addition to these methods, IW-SSIM is also added to the table because of its high performance in the LIVE and the MULTI databases. In the analysis of the LIVE and the MULTI databases, we use MSE as a fidelity method and SSIM, MS-SSIM, CW-SSIM, and IW-SSIM as structural methods. Moreover,  we also include the highest performing methods in the TID database (FSIMc and PSNR-HA). We report the results corresponding to the building blocks and variants of the proposed method to show the significance of each block in the analysis. The variants of \texttt{CSV} are $CSV_b$, $CSV_n$, and $CSV_e$, in which the subscript $b$ and $n$ correspond to interpolation methods $bilinear$ and $nearest$ $neighbor$, and $e$ corresponds to substituting the \texttt{CIEDE} formulation in Eq. \ref{eq:csv_pool_1} with the average Euclidean distance between color channels.  

\begin{center}

\begin{table}[htbp!]

  \centering
    \tiny
    \caption{Performance of the methods in the MULTI database.}

    \begin{tabular}{|c||c|c|c|}   \hline

    \multirow{2}[3]{*}{\textbf{Sequence}}

          & \textbf{Blur+JPEG}& \textbf{Blur+Noise} & \textbf{All}   
          
 \\ \cline{2-4}

    & \multicolumn{3}{c|}{\textbf{Pearson (PLCC)}}               \\ \hline

	\textbf{CSV} &\textbf{0.836}&\textbf{0.871}&\textbf{0.852}  \\ 
	\textbf{$CSV_b$} &0.000&\bf -0.001&-0.002  \\ 
	\textbf{$CSV_n$} &0.000&\bf -0.001&-0.002  \\  
	\textbf{$CSV_e$} &-0.029&-0.013&-0.024  \\ \hline
 	\textbf{CIEDE}  &-0.610&-0.212&-0.420\\ 
	\textbf{CND} &-0.493&-0.211&-0.359\\ 
	\textbf{RGCD}  &-0.057&-0.054&-0.057\\ 
	\textbf{SD}  &-0.011&-0.100&-0.074\\ 
	\textbf{RGCD$\cdot$SD} &\bf +0.015&\bf +0.015&\bf +0.016\\ \hline
	\textbf{MSE} &-0.107&-0.100&-0.110\\ 
  \textbf{SSIM} &-0.039&-0.037&-0.039\\  
  \textbf{MS-SSIM} &-0.043&-0.032&-0.049\\ 
 \textbf{IW-SSIM} &\textbf{+0.001}&-0.012&\bf 0.000\\
 \textbf{CW-SSIM} &-0.428&-0.495&-0.472\\ 
  \textbf{FSIMc} &-0.017&-0.053&-0.033\\ 
 \textbf{PSNR-HA} &\textbf{+0.001}&-0.067&-0.030\\ \hline

    & \multicolumn{3}{c|}{\textbf{Spearman}}           \\ \hline
 
    \textbf{CSV}&0.840&0.856&0.848       \\ 
	\textbf{$CSV_b$} &-0.001&-0.003&-0.002  \\ 
	\textbf{$CSV_n$} &0.000&-0.002&-0.002  \\ 
	\textbf{$CSV_e$} &-0.039&-0.015&-0.034  \\ \hline
	\textbf{CIEDE}  &-0.567&-0.256&-0.493\\ 
    \textbf{CND}  &-0.495&-0.312&-0.435\\ 
	\textbf{RGCD} &-0.062&-0.066&-0.064\\ 
	\textbf{BSD}  &-0.041&-0.135&-0.125\\ 
	\textbf{RGCD$\cdot$SD}  &+0.007&\textbf{+0.017}&+0.010\\ \hline
 	\textbf{MSE} &-0.178&-0.147&-0.171\\ 
	  \textbf{SSIM} &\textbf{+0.008}&\textbf{+0.019}&\textbf{+0.011}\\ 
  \textbf{MS-SSIM} &0.000&+0.006&-0.012\\	
 \textbf{IW-SSIM} &\textbf{+0.029}&\textbf{+0.037}&\textbf{+0.034}\\ 
  \textbf{CW-SSIM} &-0.201&-0.224&-0.218\\ 
 \textbf{FSIMc} &\textbf{+0.013}&+0.008&\textbf{+0.014}\\ 
 \textbf{PSNRHA}  &-0.098&-0.118&-0.105\\ \hline

    & \multicolumn{3}{c|}{\textbf{Kendall}}       \\ \hline    
     \textbf{CSV} &0.646&0.665&0.655       \\ 
	\textbf{$CSV_b$} &-0.002&-0.005&-0.004  \\ 
	\textbf{$CSV_n$} &-0.001&-0.005&-0.003  \\ 
	\textbf{$CSV_e$} &-0.046&-0.012&-0.036  \\   \hline 
	\textbf{CIEDE}  &-0.460&-0.226&-0.405\\ 
	\textbf{CND}   &-0.410&-0.275&-0.367\\ 
	\textbf{RGCD}   &-0.068&-0.066&-0.069\\ 
	\textbf{SD}   &-0.050&-0.137&-0.126\\ 
	\textbf{RGCD$\cdot$SD}  &\textbf{+0.006}&\textbf{+0.017}&+0.013\\ \hline
 	\textbf{MSE} &-0.169&-0.136&-0.155\\ 
	  \textbf{SSIM} &+0.005&\textbf{+0.021}&\textbf{+0.014}\\ 
  \textbf{MS-SSIM}  &-0.003&+0.009&-0.011\\ 
 \textbf{IW-SSIM} &\textbf{+0.036}&\textbf{+0.043}&\textbf{+0.046}\\ 
  \textbf{CW-SSIM} &-0.181&-0.204&-0.198\\ 
 \textbf{FSIMc} &\textbf{+0.013}&+0.004&\textbf{+0.017}\\ 
 \textbf{PSNRHA} &-0.094&-0.111&-0.096\\ \hline
    \end{tabular}%
  \label{tab:Multi_Results}
\end{table} 

\end{center}

\vspace{-12.0mm}

In the full LIVE database, \texttt{CSV} and its variants are the most linear quality estimators whereas \texttt{RGCD$\cdot$SD}, IW-SSIM, and FSIMc are the highest performing quality estimators in terms of monotonic behavior. In the Jp2k category, \texttt{CSV} and its variants lead in terms of linearity and monotonic behavior along with \texttt{RGCD$\cdot$SD} and FSIMc in terms of monotonic behavior. In the JPEG category, \texttt{CSV} and its variants lead in terms of linearity whereas SSIM, MS-SSIM, IW-SSIM, and FSIMc lead in terms of monotonic behavior. In white noise category, at least one of the \texttt{CSV} variants leads in two correlation categories along with \texttt{RGCD}, \texttt{RGCD$\cdot$SD}, MSE, and  PSNR-HA in one of the correlation types. In Gaussian blur category, \texttt{RGCD$\cdot$SD} leads in all correlation types along with \texttt{CSV} and its variant,  IW-SSIM, and FSIMc in one correlation category. In fast-fading category, \texttt{SD},  \texttt{RGCD$\cdot$SD}, SSIM, MS-SSIM, and FSIMc lead in at least one correlation type. In the MULTI database, \texttt{CSV}, \texttt{RGCD$\cdot$SD}, IW-SSIM, SSIM, and FSIMc are the best performing quality estimators in at least one correlation category. In blur+JPEG category, IW-SSIM leads in all the categories along with \texttt{CSV}, \texttt{RGCD$\cdot$SD}, SSIM, FSIMc, and PSNR-HA in one of the correlation categories. In blur+noise category,  \texttt{RGCD$\cdot$SD} is among the highest performing quality estimators in all three correlation categories along with \texttt{CSV} and its variants, SSIM, and IW-SSIM.

In the TID database, PSNR-HA, PSNR-HMA, and PSNR-HVS are the highest performing quality estimators in the noise and the actual categories. In the simple category, PSNR-HA, PSNR-HVS, and FSIM have the best monotonic behavior. In the exotic category, FSIM, FSIMc, and MS-SSIM are the highest performing quality estimators in both correlation types and IW-SSIM is also among the best performers in Spearman correlation category. In the new and the color categories, \texttt{CSV} and its variants are the best performing quality estimators in terms of both correlation categories. In the full TID database, \texttt{CSV}, $CSV_b$, and FSIMc are the best performing quality estimators. When we consider the LIVE, the MULTI, and the TID databases, \texttt{CSV} is the best method compared to the building blocks of \texttt{CSV}. \texttt{RGCD$\cdot$SD} performs as good as \texttt{CSV} in the LIVE and the MULTI databases. However, \texttt{CSV} outperforms \texttt{RGCD$\cdot$SD} in the TID database.

\begin{center}
\begin{table*}[htbp!]
\begin{adjustwidth}{-3.5cm}{}

  \centering
    \tiny
    \caption{Performance of the methods in the TID 2013 database.}
    \begin{tabular}{|c||c|c|c|c|c|c|c||c|c|c|c|c|c|c|}   \hline


    \multirow{2}[3]{*}{\textbf{Sequence}} 
    
    & \multicolumn{7}{c||}{\textbf{Spearman (SCC)}}     & \multicolumn{7}{c|}{\textbf{Kendall (KCC)}}  \\ \cline{2-15}

          & \textbf{Noise}& \textbf{Actual}& \textbf{Simple}& \textbf{Exotic}& \textbf{New}  & \textbf{Color} &\textbf{Full}  
          & \textbf{Noise}& \textbf{Actual}& \textbf{Simple}& \textbf{Exotic}& \textbf{New}  & \textbf{Color} &\textbf{Full}

          \\ \hline

	\textbf{CSV} &0.849 &0.880 &0.924 &0.812 &\textbf{0.898}  &\textbf{0.888}  &\textbf{0.845}
	 &0.665 &0.710 &0.758 &0.605 &\textbf{0.724}  &\textbf{0.713}  &\textbf{0.654}  \\ 

		\textbf{$CSV_b$} &+0.006&+0.003&-0.003&-0.003&\bf +0.013&\bf +0.021&\bf +0.003 
		&+0.006&+0.003&-0.007&-0.004&\bf +0.016&\bf +0.028&\bf +0.004 \\ 
		\textbf{$CSV_n$} &+0.007&+0.004&-0.003&-0.006&+\bf 0.012&\bf +0.021&+0.002 
		&+0.007&+0.004&-0.008&-0.007&\bf +0.015&\bf +0.028&+0.002 \\ 
		\textbf{$CSV_e$} &-0.008&0.000&+0.012&-0.042&-0.049&-0.081&-0.048
		 &-0.008&0.000&+0.021&-0.044&-0.053&-0.093&-0.049 \\ \hline
		\textbf{CIEDE} &-0.019&-0.016&-0.115&-0.416&-0.145&-0.026&-0.229
	     &-0.032&-0.034&-0.152&-0.337&-0.181&-0.058&-0.211\\ 
	\textbf{CND} &-0.068&-0.066&-0.223&-0.505&-0.136&-0.044&-0.267
	&-0.082&-0.088&-0.251&-0.381&-0.164&-0.071&-0.237\\ 
	\textbf{RGCD} &+0.023&+0.011&+0.009&-0.089&+0.005&-0.005&-0.030
	&+0.023&+0.005&+0.020&-0.053&+0.010&-0.009&-0.015\\ 
	\textbf{SD} &-0.287&-0.225&-0.174&-0.227&-0.289&-0.370&-0.301
	&-0.277&-0.244&-0.217&-0.191&-0.285&-0.348&-0.267\\ 
	\textbf{RGCD$\cdot$SD}&-0.024&-0.017&-0.001&-0.005&-0.101&-0.138&-0.040
	&-0.024&-0.021&0.000&+0.010&-0.105&-0.141&-0.032\\ \hline
	\textbf{FSIMc} &+0.052&+0.034&+0.022&\textbf{+0.028}&-0.110&-0.113&\textbf{+0.005}
	&+0.057&+0.032&+0.034&\textbf{+0.046}&-0.112&-0.121&\textbf{+0.012}\\ 
    \textbf{PSNR-HA} &\textbf{+0.073}&\textbf{+0.057}&\textbf{+0.027}&+0.012&-0.197&-0.256&-0.026
    &\textbf{+0.094}&\textbf{+0.076}&\textbf{+0.059}&+0.018&-0.182&-0.236&-0.010\\ 
    \textbf{PSNR-HMA} &\textbf{+0.065}&\textbf{+0.053}&+0.012&+0.001&-0.160&-0.214&-0.032
    &\textbf{+0.079}&\textbf{+0.066}&+0.026&+0.004&-0.151&-0.206&-0.022\\ 
	\textbf{FSIM} &+0.047&+0.030&\textbf{+0.023}&\textbf{+0.031}&-0.249&-0.323&-0.044
	&+0.050&+0.026&\textbf{+0.036}&\textbf{+0.049}&-0.200&-0.261&-0.024\\ 
	\textbf{MS-SSIM} &+0.023&+0.006&-0.019&\textbf{+0.029}&-0.267&-0.322&-0.058
	&+0.014&-0.012&-0.037&\textbf{+0.042}&-0.229&-0.258&-0.046\\ 
	\textbf{IW-SSIM}
&+0.021&+0.006&-0.013&\bf +0.028&-0.279&-0.339&-0.067
&+0.013&-0.009&-0.028&+0.038&-0.248&-0.289&-0.056\\ 
	\textbf{PSNRc} &-0.080&-0.077&-0.049&-0.249&-0.121&-0.154&-0.158
	&-0.103&-0.114&-0.069&-0.213&-0.148&-0.178&-0.158\\ 
	\textbf{VSNR} &+0.019&+0.001&-0.012&-0.105&-0.309&-0.376&-0.164
	&+0.010&-0.019&-0.027&-0.086&-0.286&-0.335&-0.146\\ 
	\textbf{PSNR-HVS} &\textbf{+0.067}&\textbf{+0.045}&\textbf{+0.025}&-0.211&-0.251&-0.329&-0.191
	&\textbf{+0.089}&\textbf{+0.055}&\textbf{+0.050}&-0.170&-0.207&-0.265&-0.146\\ 
	\textbf{PSNR} &-0.028&-0.055&-0.011&-0.215&-0.279&-0.349&-0.206
	&-0.042&-0.086&-0.013&-0.180&-0.251&-0.298&-0.184\\ 
	\textbf{SSIM} &-0.092&-0.092&-0.087&-0.180&-0.318&-0.382&-0.208
	&-0.114&-0.133&-0.130&-0.150&-0.301&-0.331&-0.190 \\ 
	\textbf{NQM} &-0.013&-0.023&-0.049&-0.222&-0.272&-0.346&-0.210
	&-0.024&-0.044&-0.077&-0.193&-0.241&-0.305&-0.188\\ 
	\textbf{PSNR-HVS-M} &+0.056&+0.037&+0.013&-0.247&-0.251&-0.331&-0.220
	&+0.067&+0.038&+0.021&-0.202&-0.206&-0.273&-0.172\\ 
	\textbf{VIFP} &-0.066&-0.065&-0.027&-0.254&-0.306&-0.379&-0.237
	&-0.078&-0.089&-0.044&-0.199&-0.273&-0.320&-0.197\\ 
	\textbf{WSNR}&+0.030&+0.016&+0.008&-0.389&-0.251&-0.329&-0.265
	&+0.030&+0.007&+0.014&-0.308&-0.209&-0.277&-0.207\\ \hline
    \end{tabular}%
  \label{tab:TID_Results}
\vspace{-1.0mm}
\end{adjustwidth}{}
\end{table*}

\end{center}

When we consider complete databases, the best performing quality estimators that are highlighted in the tables are SSIM, IW-SSIM, FSIMc, \texttt{CSV}, its variants and building blocks. The selection of the interpolation strategy only slightly affects the quality estimation performance. In the LIVE and the MULTI databases, \texttt{CSV} outperforms $CSV_b$ and $CSV_n$ whereas in the TID database, $CSV_b$ outperforms $CSV_n$, and $CSV_n$ outperforms \texttt{CSV}. \texttt{CSV} outperforms $CSV_e$ in the overall databases in all the correlation categories. To analyze the performance difference between best performing quality estimators, we perform the statistical significance tests suggested in ITU-T Rec. P.1401. \cite{ITU-T}.

\begin{center}

\begin{table*}[htbp!]                                                                                                                                    
\centering                                                                                                                                           
\tiny
\begin{tabular}{|c|c|c|c|c|}                                                                                                                       
\hline                                                                                                                                               
 & CSV & FSIMc & IWSSIM  & SSIM \\                                                                                                           
\hline                                                                                                                                               
CSV & 0  0  0  0  0  0  0  0  0 & 1  0  0  1  0  0  0  0  0 & 1  0  0  1  1  1  0  1  0  & 1  1  0  1  1  1  0  0  0 \\ \hline
FSIMc & 1  0  0  1  0  0  0  0  0 & 0  0  0  0  0  0  0  0  0 & 0  0  0  0  1  1  0  0  0  & 0  1  0  1  1  1  0  0  0 \\                                                                                                                                                
IWSSIM & 1  0  0  1  1  1  0  1  0 & 0  0  0  0  1  1  0  0  0 & 0  0  0  0  0  0  0  0  0  & 0  1  0  1  1  1  0  0  0 \\                                                                                                                                               SSIM & 1  1  0  1  1  1  0  0  0 & 0  1  0  1  1  1  0  0  0 & 0  1  0  1  1  1  0  0  0  & 0  0  0  0  0  0  0  0  0 \\  
\hline                                                                                                                                               
\end{tabular}                                                                                                                                        
\caption{ITU-T Rec. P.1401 statistical significance test results.}                                                                                    
\label{tab:sign}                                                                                                                                                                                                                                                                                                                                                                                                                                                                                                                                                                                                   
\vspace{-10.0mm}
\end{table*} 
\end{center}

\begin{center}

\begin{figure}[htbp!]

\begin{minipage}[b]{0.28\linewidth}
  \centering
\includegraphics[width=0.9\linewidth, trim= 20mm 65mm 20mm 65mm]{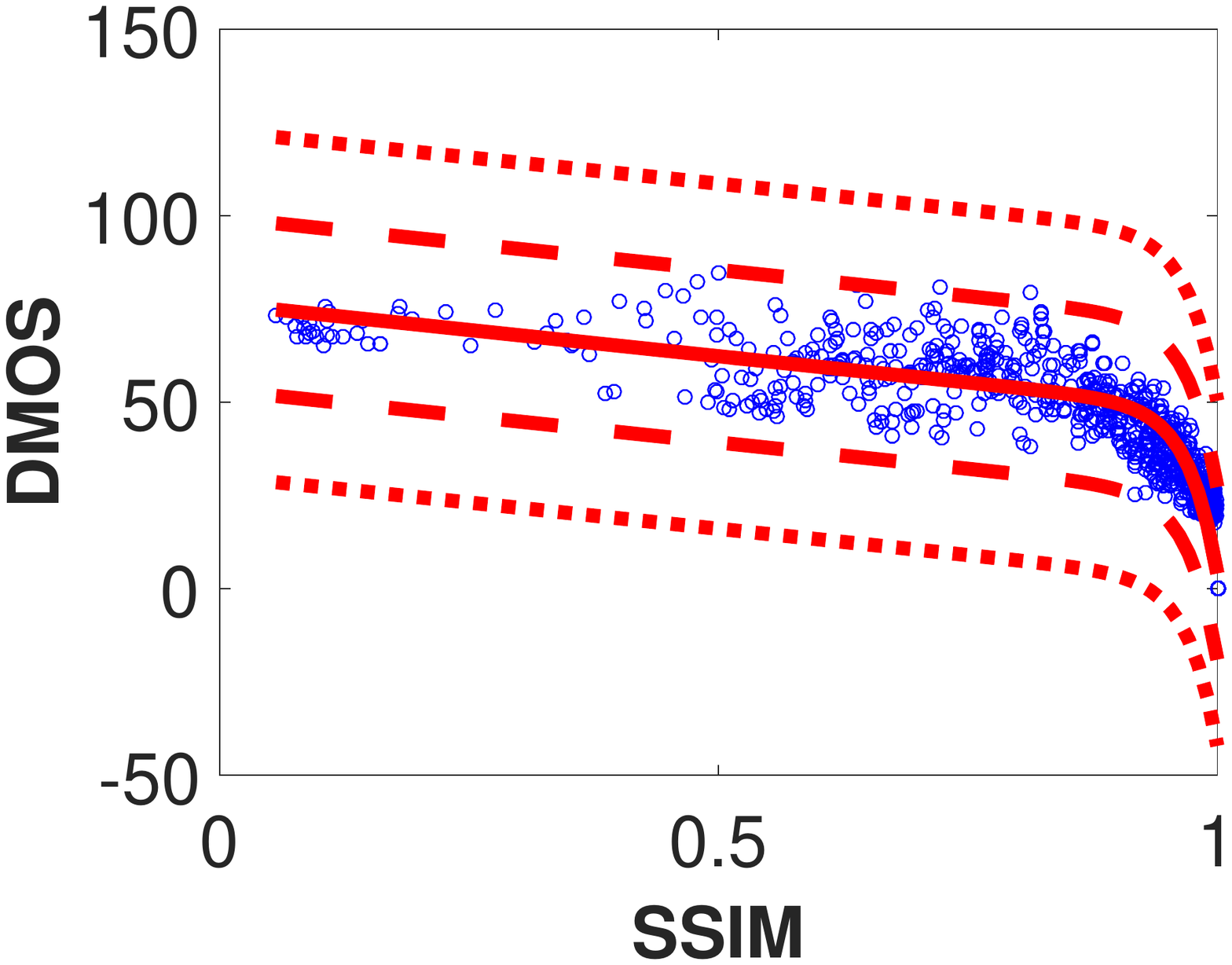}
  \vspace{0.03cm}
  \centerline{\footnotesize{(a)LIVE-SSIM}}
  \vspace{-0.45cm}

\end{minipage}
 \vspace{0.2cm}
\hfill
\begin{minipage}[b]{0.28\linewidth}
  \centering
\includegraphics[width=0.9\linewidth, trim= 20mm 65mm 20mm 65mm]{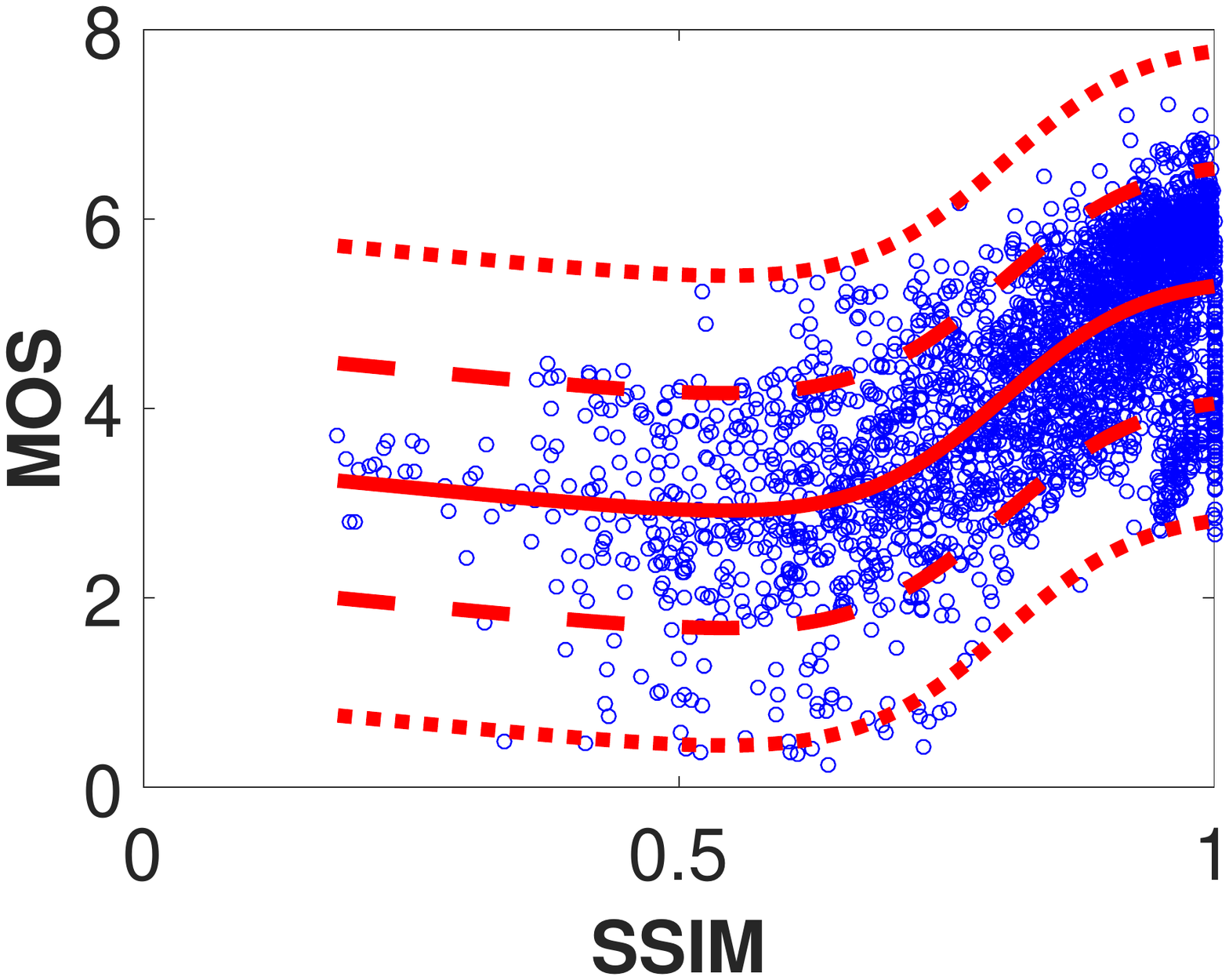}
  \vspace{0.03cm}
  \centerline{\footnotesize{(b) TID-SSIM }}
    \vspace{-0.45cm}

\end{minipage}
 \vspace{0.2cm}
\hfill
\begin{minipage}[b]{0.28\linewidth}
  \centering
\includegraphics[width=0.9\linewidth, trim= 20mm 65mm 20mm 65mm]{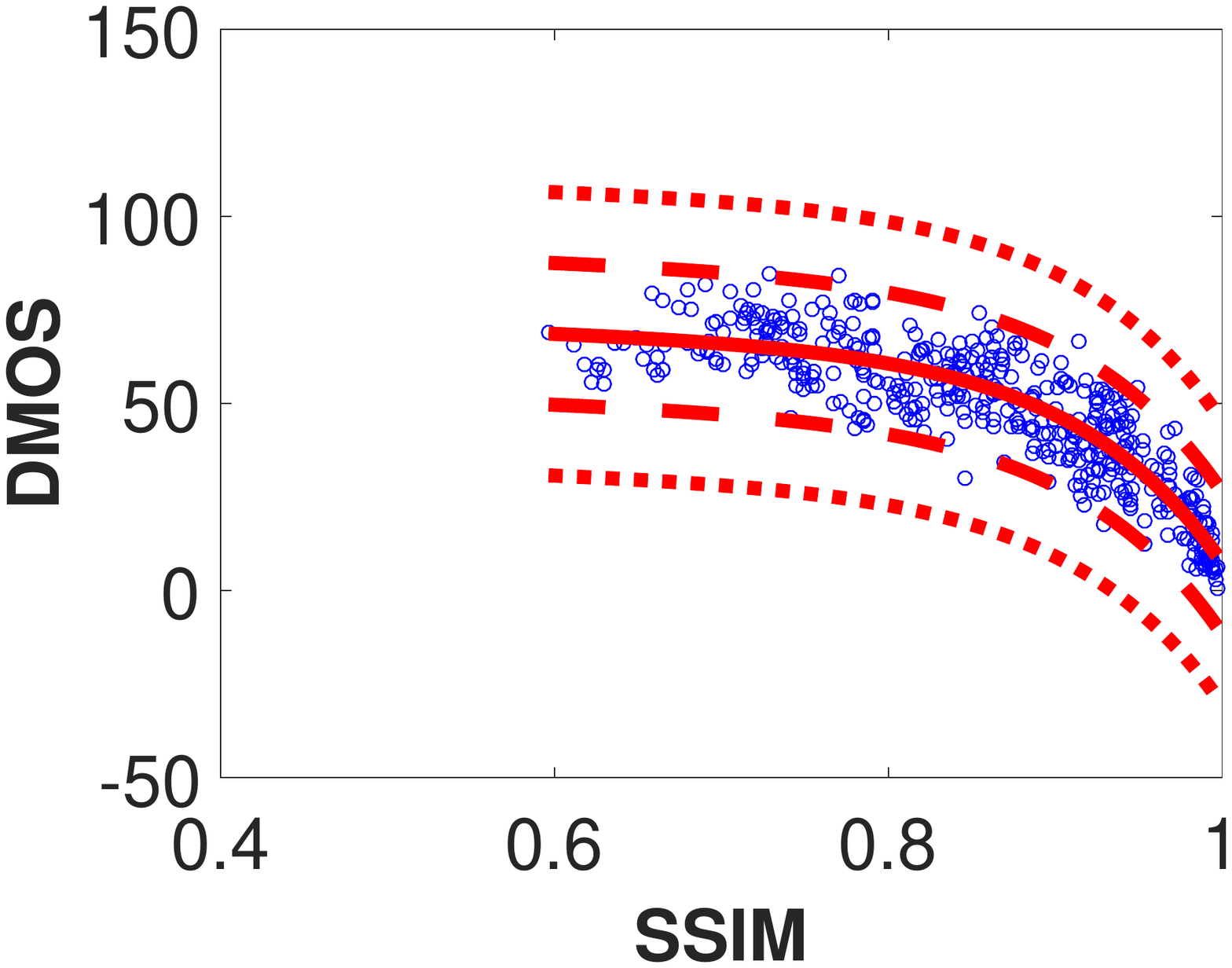}
  \vspace{0.03 cm}
  \centerline{\footnotesize{(c) MULTI-SSIM   } }
    \vspace{-0.45cm}
\end{minipage}

\begin{minipage}[b]{0.28\linewidth}
  \centering
\includegraphics[width=0.9\linewidth, trim= 20mm 65mm 20mm 65mm]{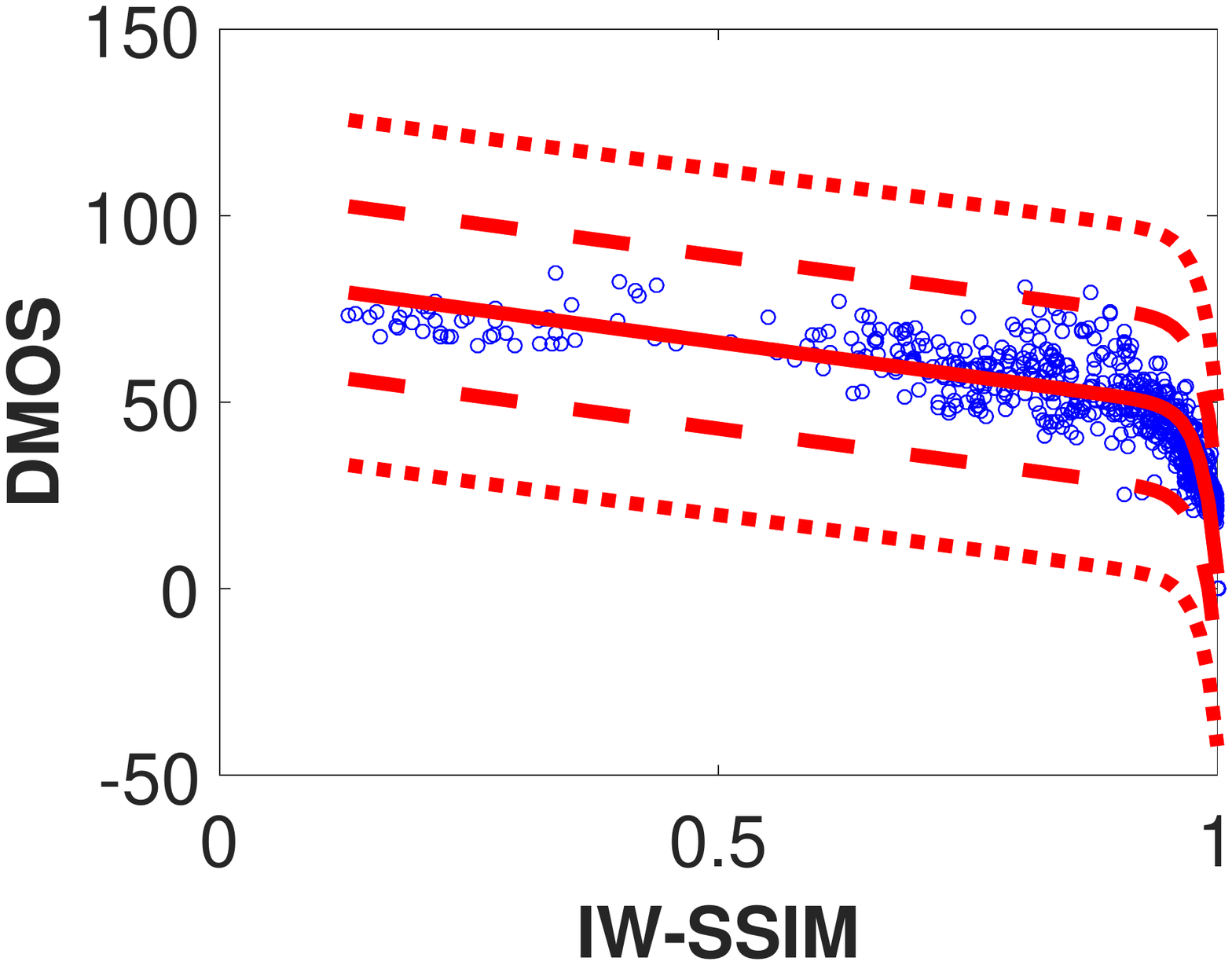}
  \vspace{0.03cm}
  \centerline{\footnotesize{(d)LIVE-IW-SSIM}}
      \vspace{-0.45cm}
\end{minipage}
 \vspace{0.2cm}
\hfill
\begin{minipage}[b]{0.28\linewidth}
  \centering
\includegraphics[width=0.9\linewidth, trim= 20mm 65mm 20mm 65mm]{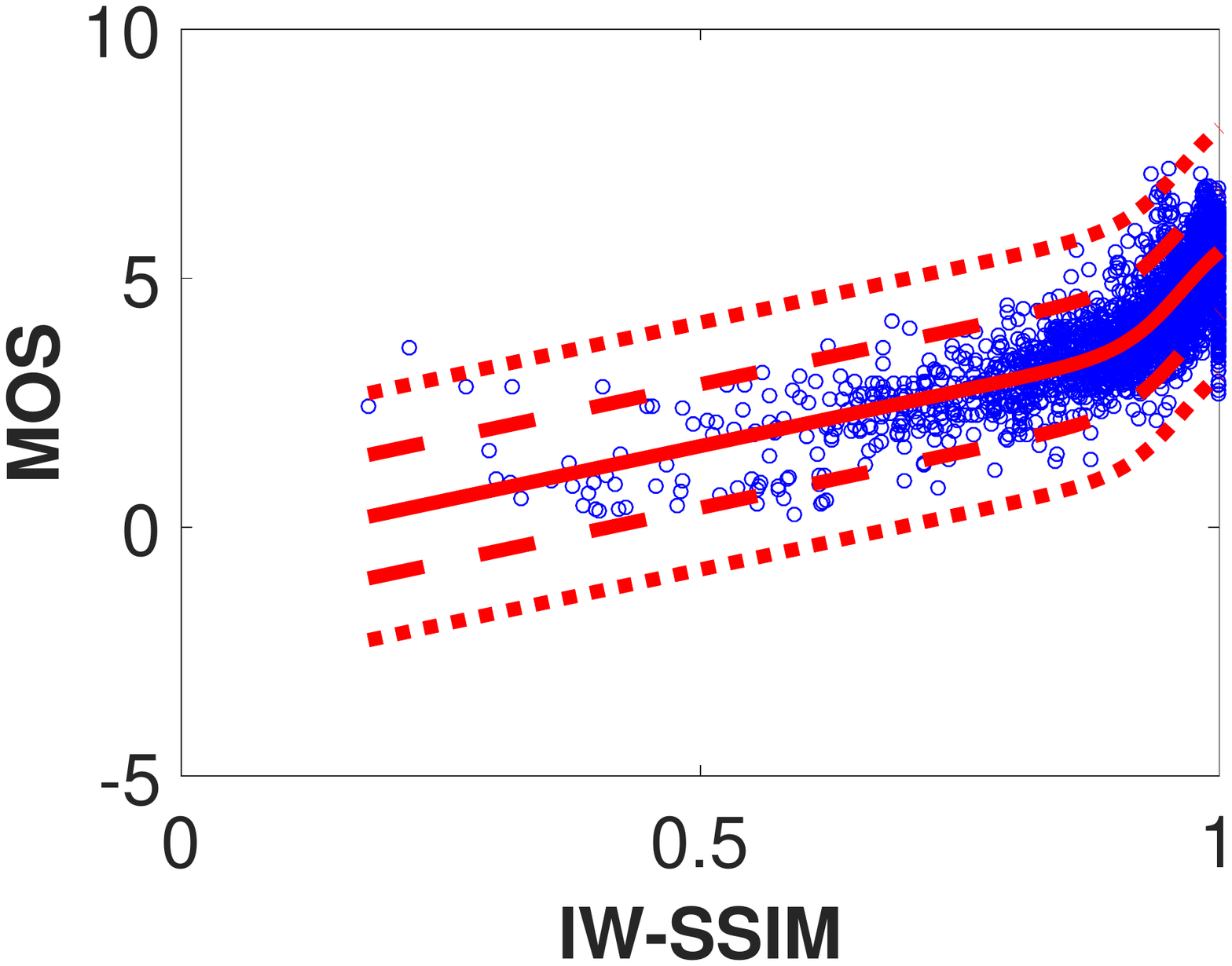}
  \vspace{0.03cm}
  \centerline{\footnotesize{(e) TID-IW-SSIM }}
      \vspace{-0.45cm}
\end{minipage}
 \vspace{0.2cm}
\hfill
\begin{minipage}[b]{0.28\linewidth}
  \centering
\includegraphics[width=0.9\linewidth, trim= 20mm 65mm 20mm 65mm]{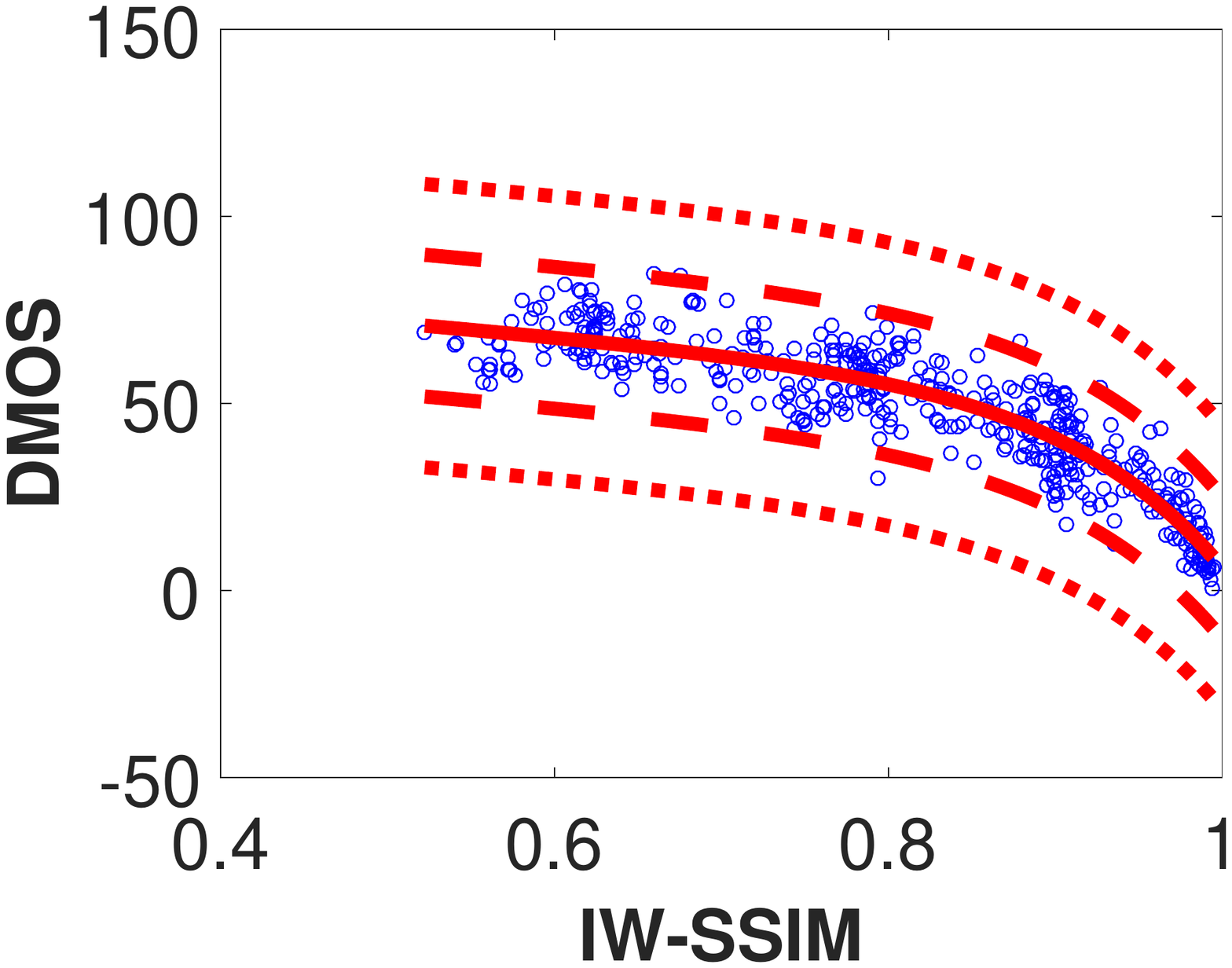}
  \vspace{0.03 cm}
  \centerline{\footnotesize{(f) MULTI-IW-SSIM   } }
      \vspace{-0.45cm}
\end{minipage}

\begin{minipage}[b]{0.28\linewidth}
  \centering
\includegraphics[width=0.9\linewidth, trim= 20mm 65mm 20mm 65mm]{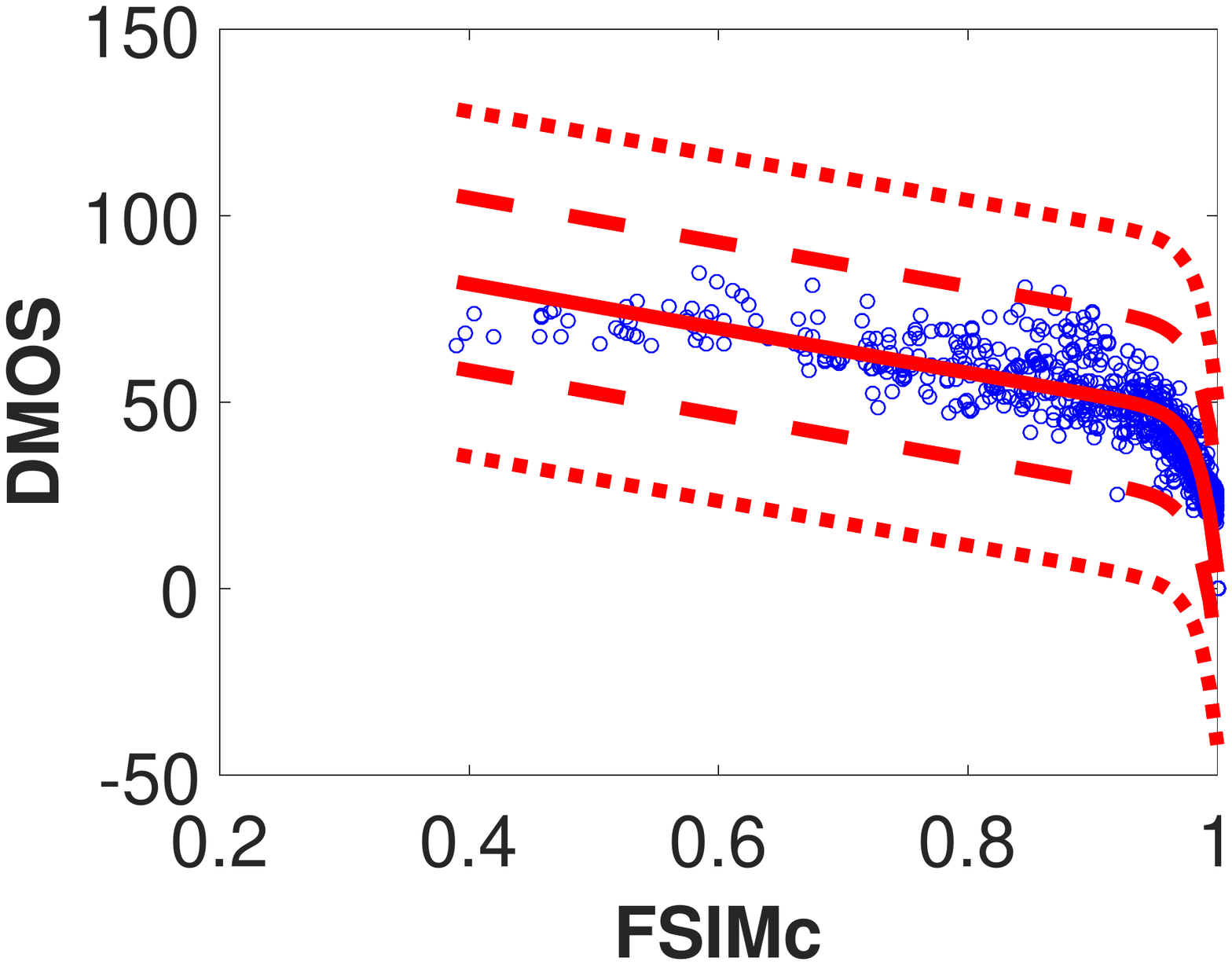}
  \vspace{0.03cm}
  \centerline{\footnotesize{(g)LIVE-FSIMc}}
      \vspace{-0.45cm}
\end{minipage}
 \vspace{0.2cm}
\hfill
\begin{minipage}[b]{0.28\linewidth}
  \centering
\includegraphics[width=0.9\linewidth, trim= 20mm 65mm 20mm 65mm]{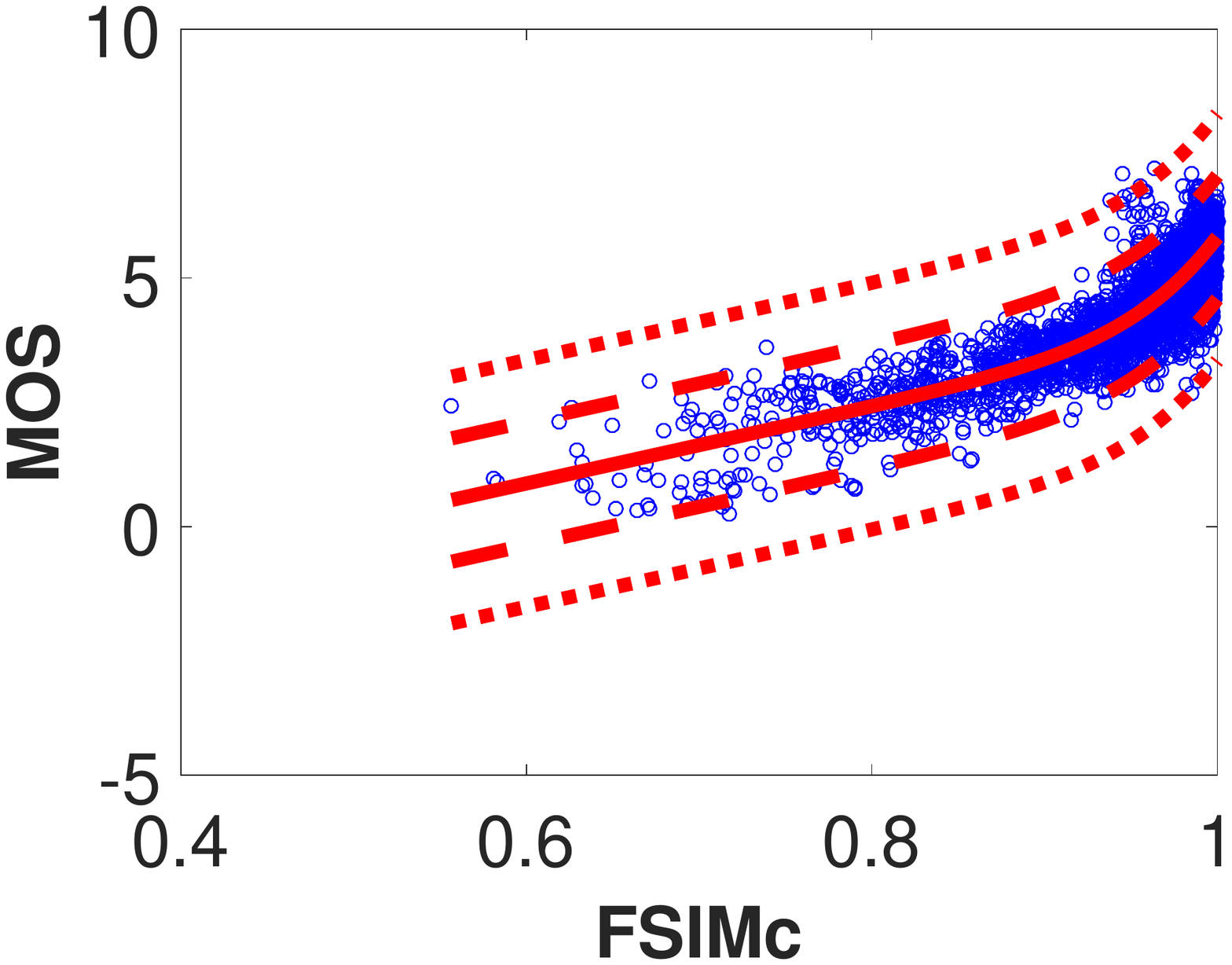}
  \vspace{0.03cm}
  \centerline{\footnotesize{(h) TID-FSIMc }}
      \vspace{-0.45cm}
\end{minipage}
 \vspace{0.2cm}
\hfill
\begin{minipage}[b]{0.28\linewidth}
  \centering
\includegraphics[width=0.9\linewidth, trim= 20mm 65mm 20mm 65mm]{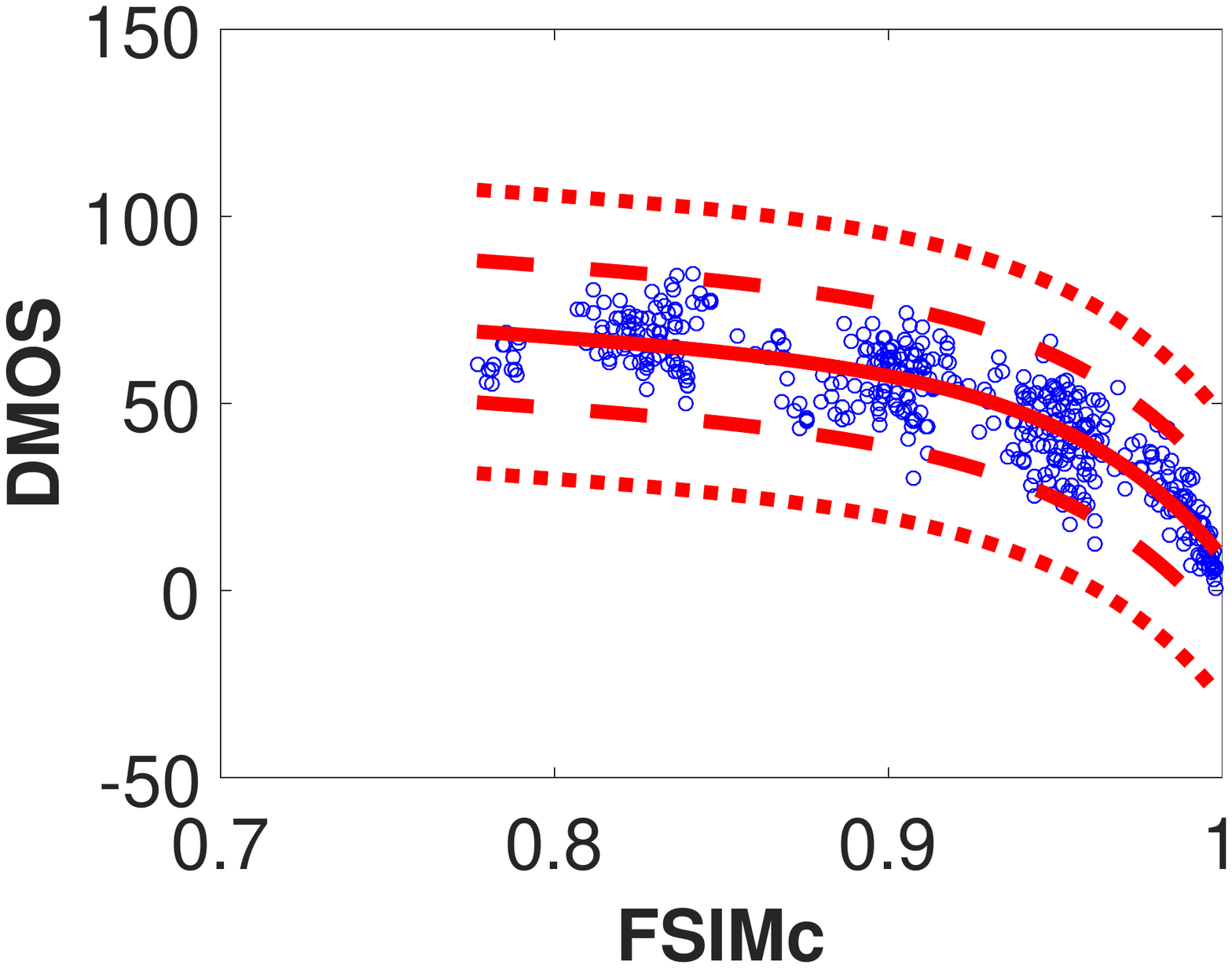}
  \vspace{0.03 cm}
  \centerline{\footnotesize{(i) MULTI-FSIMc   } }
      \vspace{-0.45cm}
\end{minipage}

\begin{minipage}[b]{0.28\linewidth}
  \centering
\includegraphics[width=0.9\linewidth, trim= 20mm 65mm 20mm 65mm]{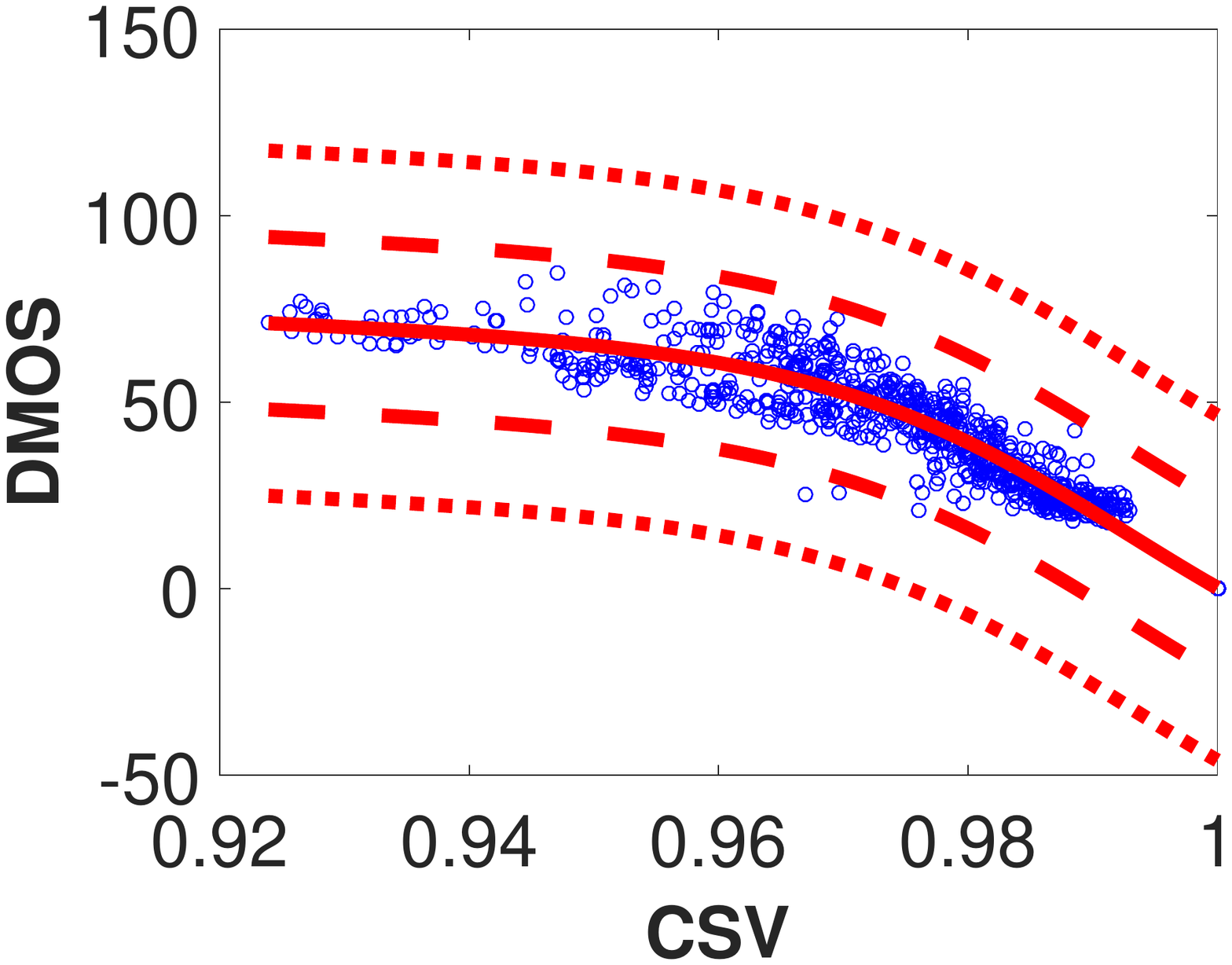}
  \vspace{0.03cm}
  \centerline{\footnotesize{(j)LIVE-CSV}}
      \vspace{-0.45cm}
\end{minipage}
 \vspace{0.2cm}
\hfill
\begin{minipage}[b]{0.28\linewidth}
  \centering
\includegraphics[width=0.9\linewidth, trim= 20mm 65mm 20mm 65mm]{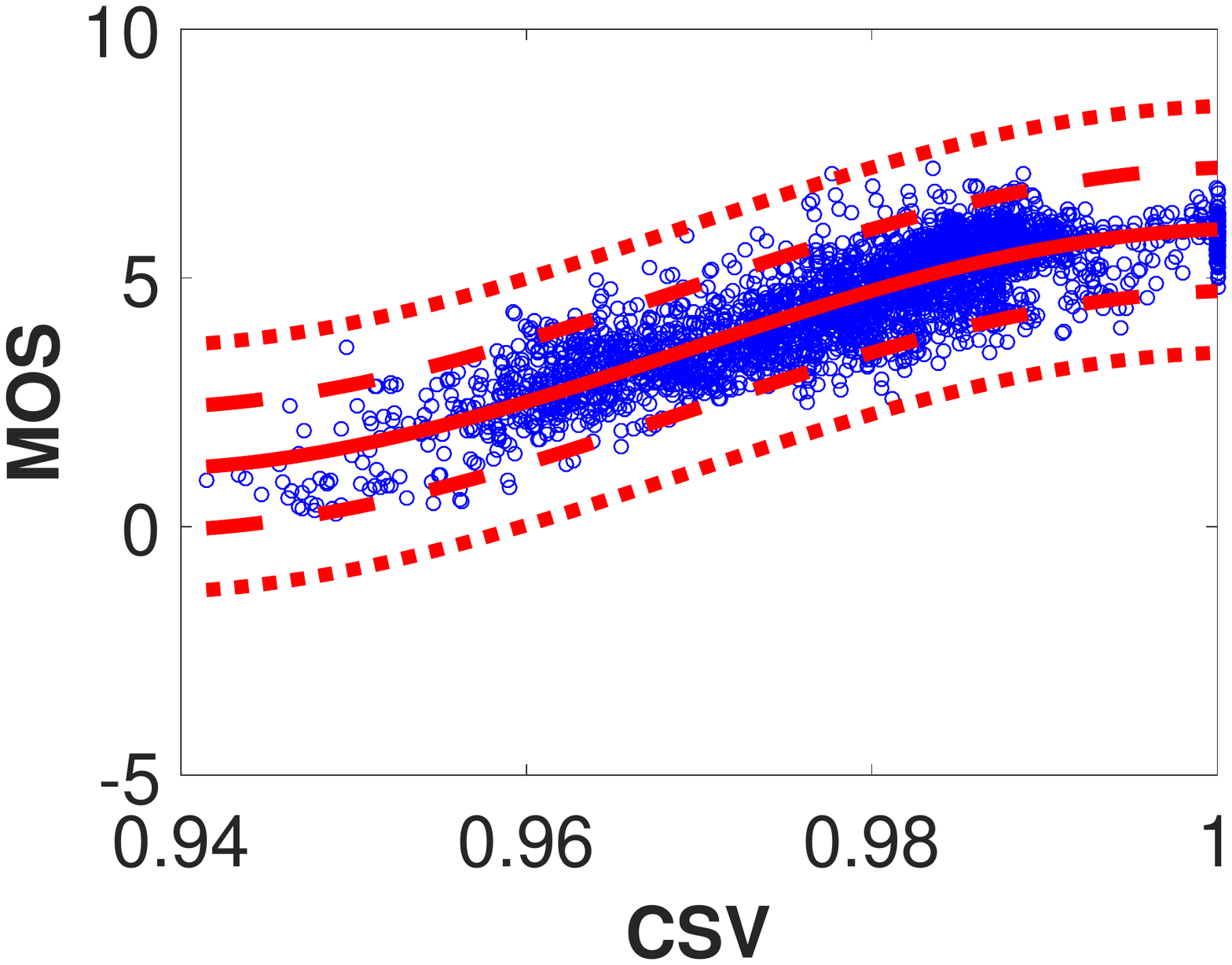}
  \vspace{0.03cm}
  \centerline{\footnotesize{(k) TID-CSV }}
      \vspace{-0.45cm}
\end{minipage}
 \vspace{0.2cm}
\hfill
\begin{minipage}[b]{0.28\linewidth}
  \centering
\includegraphics[width=0.9\linewidth, trim= 20mm 65mm 20mm 65mm]{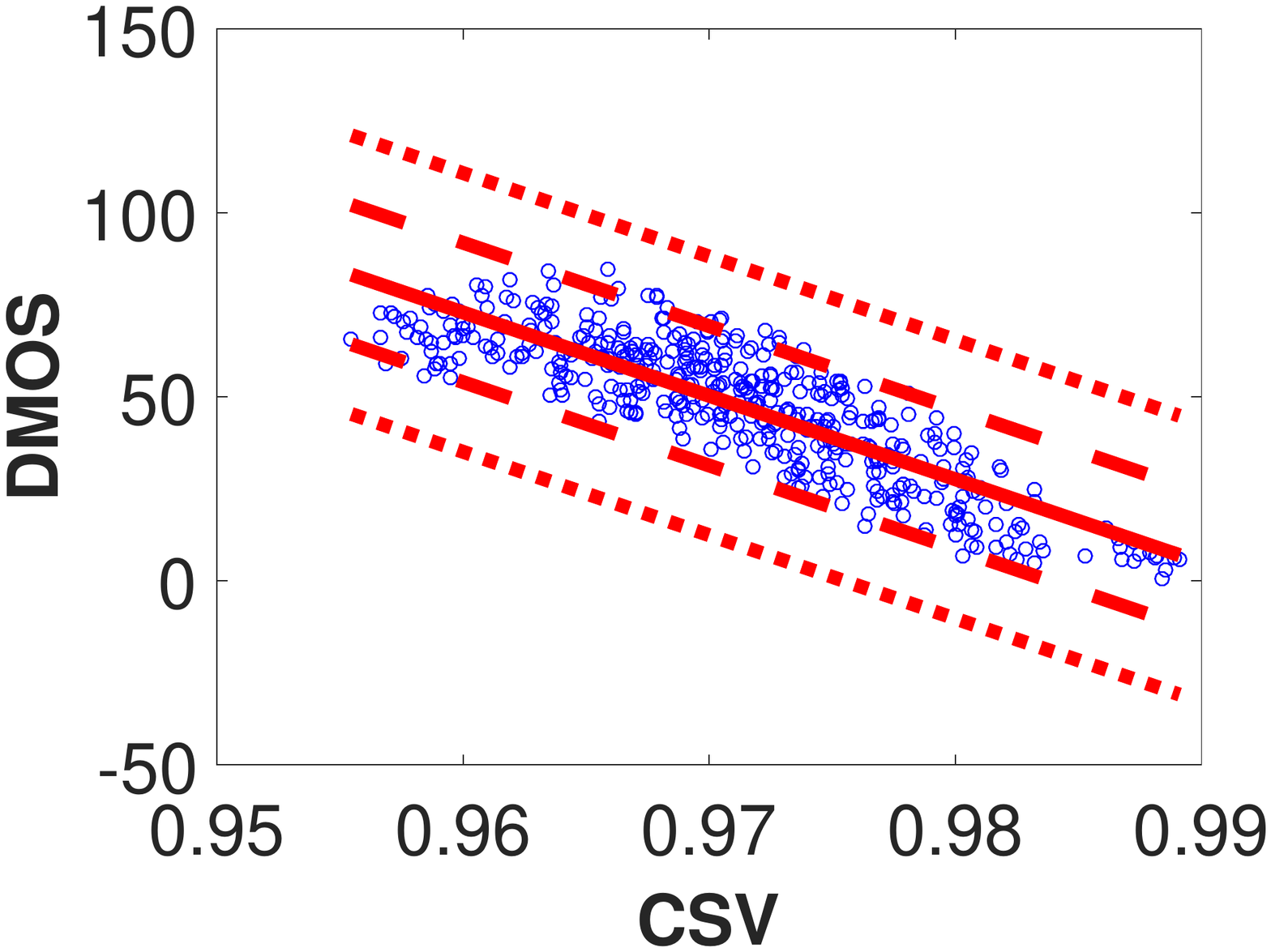}
  \vspace{0.03 cm}
  \centerline{\footnotesize{(l) MULTI-CSV   } }
      \vspace{-0.45cm}
\end{minipage}

\caption{Scatter plots of the objective quality estimates and the subjective scores.}\vspace{-.5cm}
\label{fig:Scatter}
\vspace{-4.0mm}
\end{figure}

\end{center}

Table \ref{tab:sign}  summarizes the significance of the difference between the quality assessment methods with respect to correlation coefficients. Each entry is a nine digit codeword in which the first three corresponds to the LIVE database, the second three to the TID database, and the third three to the MULTI database. In these ternary groups, the first entry corresponds to the Pearson, the second to the Spearman, and the third to the Kendall correlation coefficients. A $1$ in a codeword means that there is a significant difference between the correlation coefficients. The proposed method \texttt{CSV} is significantly different from other methods in terms of at least Pearson, Spearman or Kendall in the LIVE and the TID databases. In the MULTI database, \texttt{CSV} behaves similar to FSIMc and SSIM but significantly different from IW-SSIM in terms of Spearman correlation. For a more comprehensive comparison among these methods, we can also analyze the scatter plots of the quality estimates that are given in Fig. \ref{fig:Scatter}.

\begin{center}

\begin{figure}[htbp!]
\vspace{2.0mm}
\begin{minipage}[b]{0.28\linewidth}
  \centering
\includegraphics[width=0.8\linewidth, trim= 25mm 80mm 25mm 80mm]{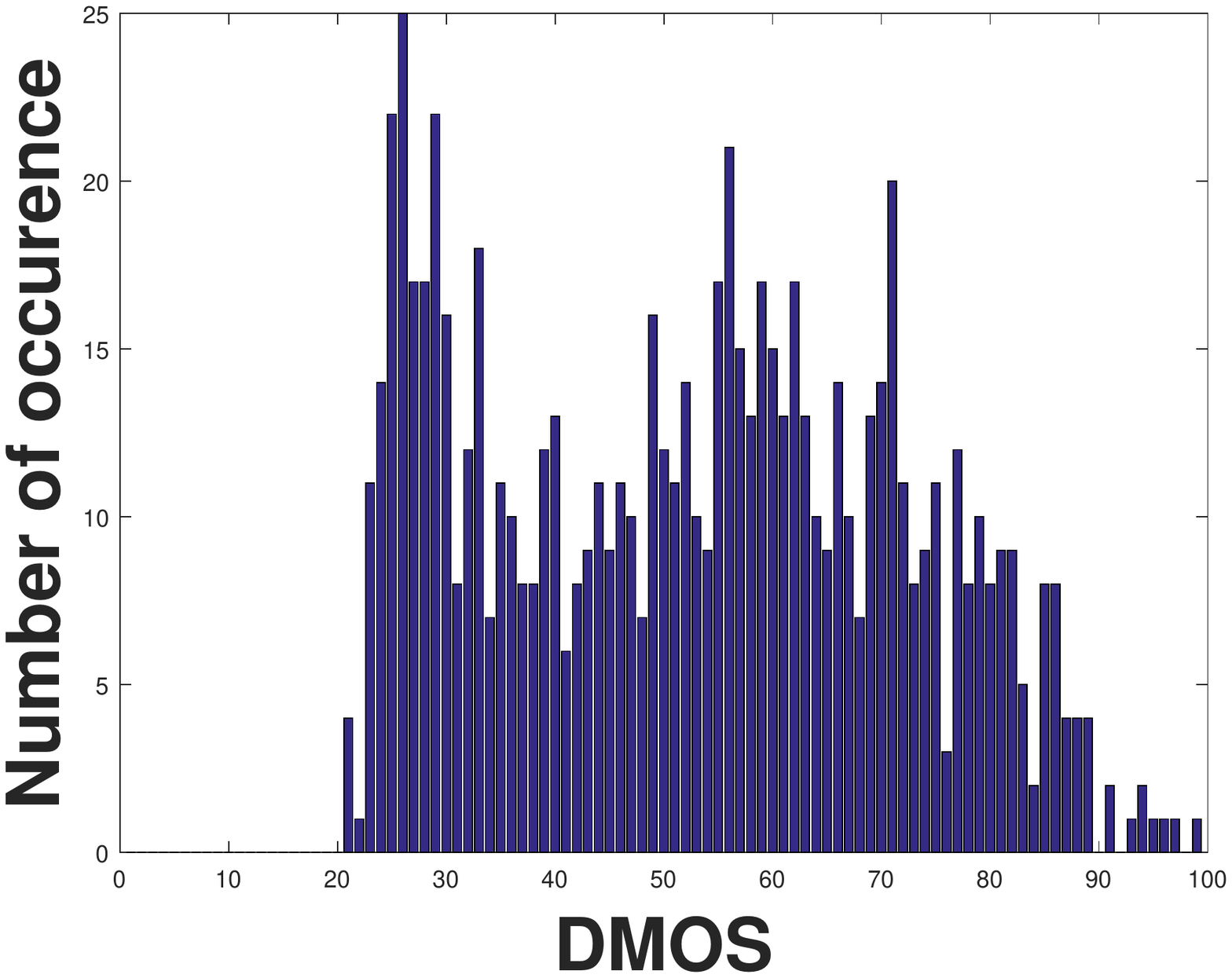}
  \vspace{0.03cm}
  \centerline{\footnotesize{(a)LIVE-DMOS}}
      \vspace{-0.20cm}

\end{minipage}
  \vspace{0.20cm}
\hfill
\begin{minipage}[b]{0.28\linewidth}
  \centering
\includegraphics[width=0.8\linewidth, trim= 25mm 80mm 25mm 80mm]{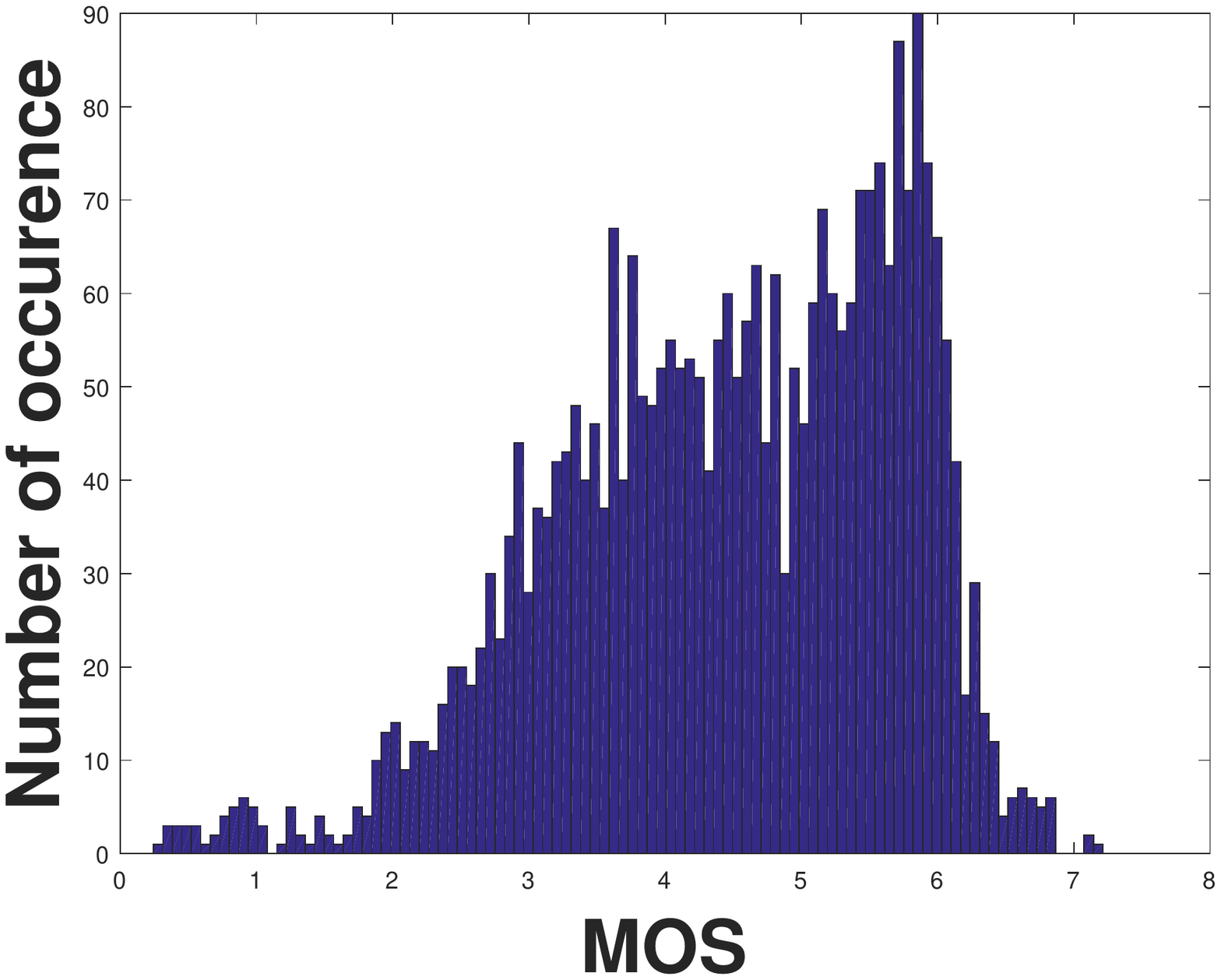}
  \vspace{0.03cm}
  \centerline{\footnotesize{(b) TID-MOS }}
      \vspace{-0.20cm}

\end{minipage}
  \vspace{0.20cm}
\hfill
\begin{minipage}[b]{0.28\linewidth}
  \centering
\includegraphics[width=0.8\linewidth, trim= 25mm 80mm 25mm 80mm]{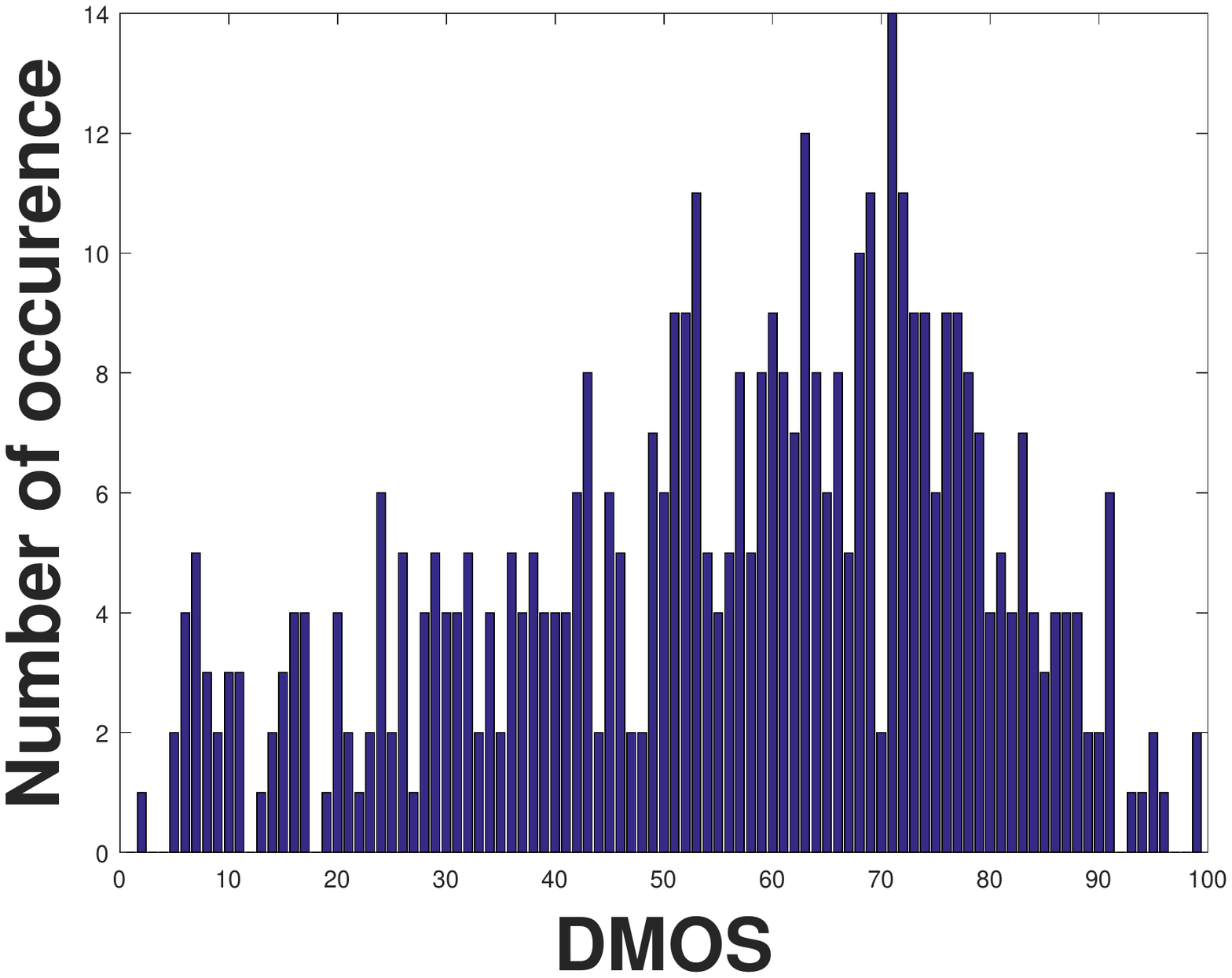}
  \vspace{0.03cm}
  \centerline{\footnotesize{(c) MULTI-DMOS }}
      \vspace{-0.20cm}

\end{minipage}

\begin{minipage}[b]{0.28\linewidth}
  \centering
\includegraphics[width=0.8\linewidth, trim= 25mm 80mm 25mm 80mm]{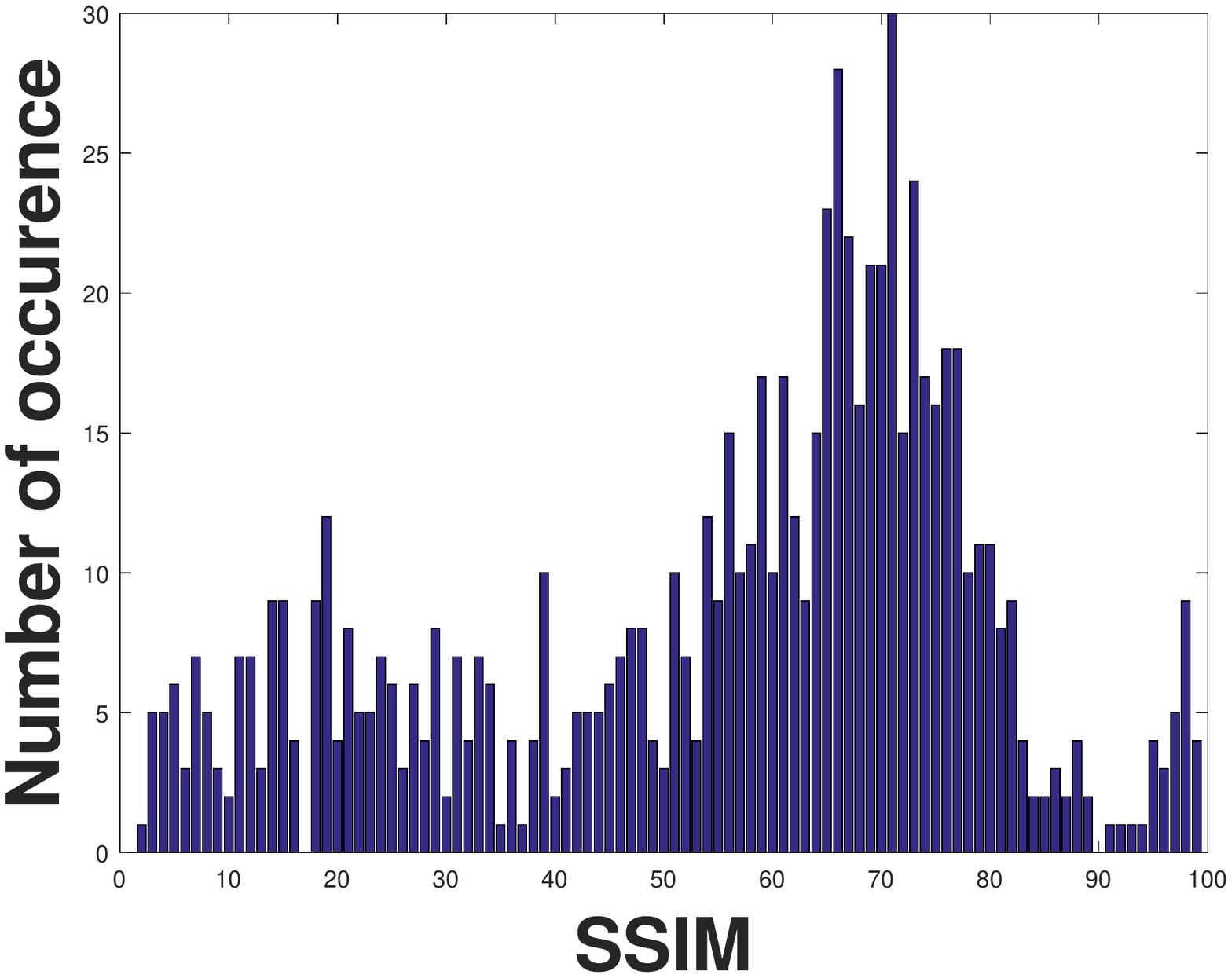}
  \vspace{0.03cm}
  \centerline{\footnotesize{(d)LIVE-SSIM}}
      \vspace{-0.20cm}
\end{minipage}
  \vspace{0.20cm}
\hfill
\begin{minipage}[b]{0.28\linewidth}
  \centering
\includegraphics[width=0.8\linewidth, trim= 25mm 80mm 25mm 80mm]{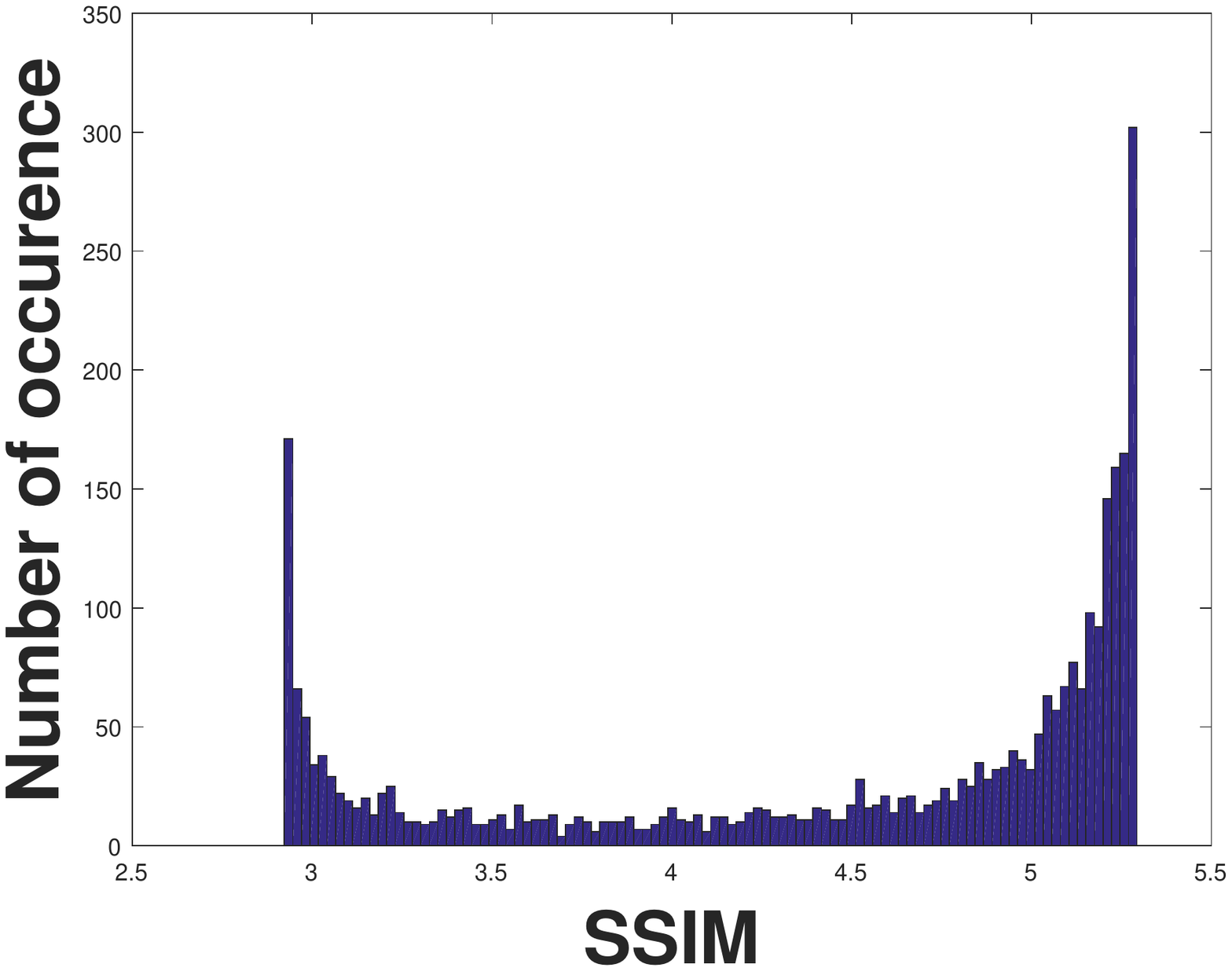}
  \vspace{0.03cm}
  \centerline{\footnotesize{(e) TID-SSIM }}
      \vspace{-0.20cm}
\end{minipage}
  \vspace{0.20cm}
\hfill
\begin{minipage}[b]{0.28\linewidth}
  \centering
\includegraphics[width=0.8\linewidth, trim= 25mm 80mm 25mm 80mm]{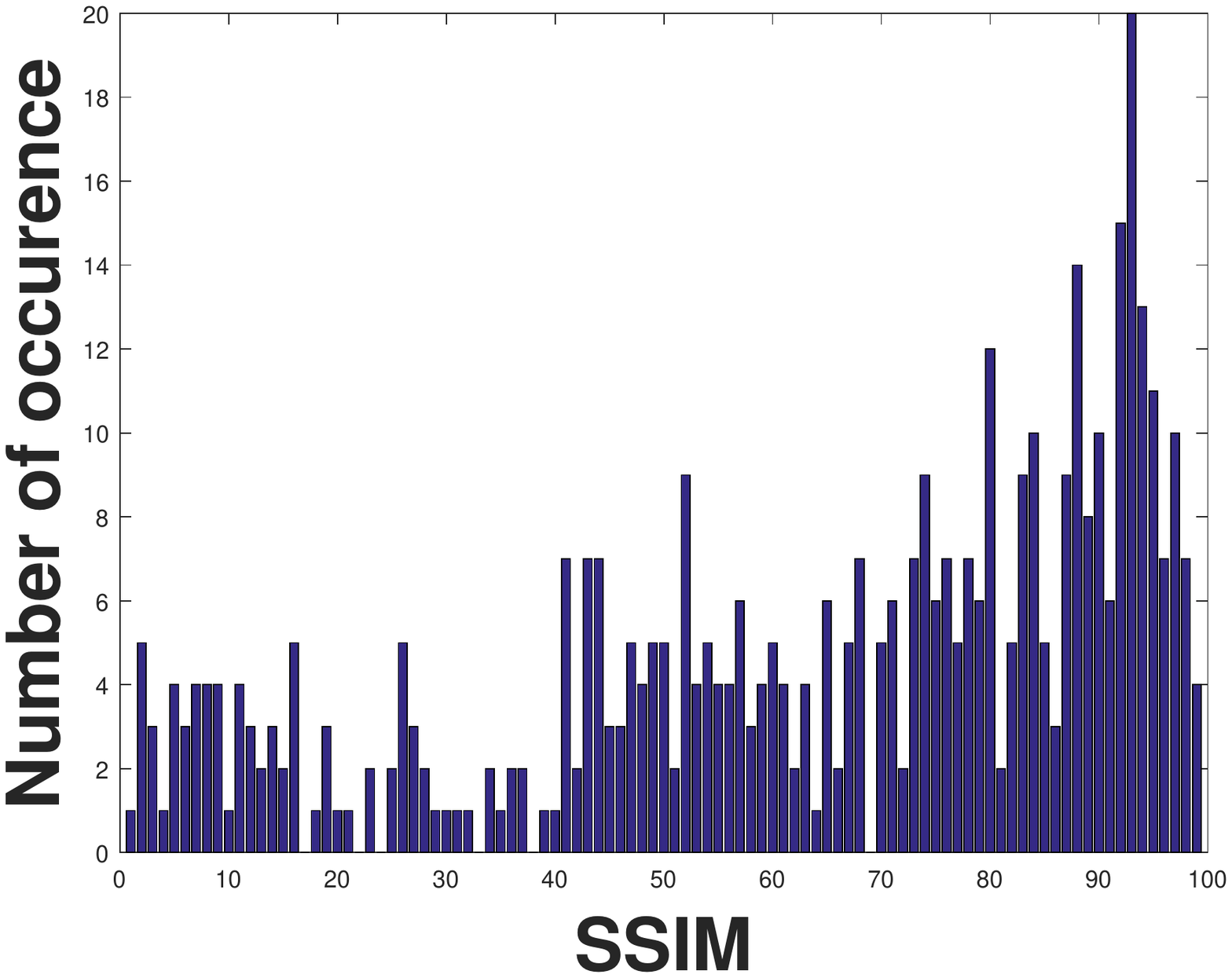}
  \vspace{0.03cm}
  \centerline{\footnotesize{(f) MULTI-SSIM }}
      \vspace{-0.20cm}
\end{minipage}

\begin{minipage}[b]{0.28\linewidth}
  \centering
\includegraphics[width=0.8\linewidth, trim= 25mm 80mm 25mm 80mm]{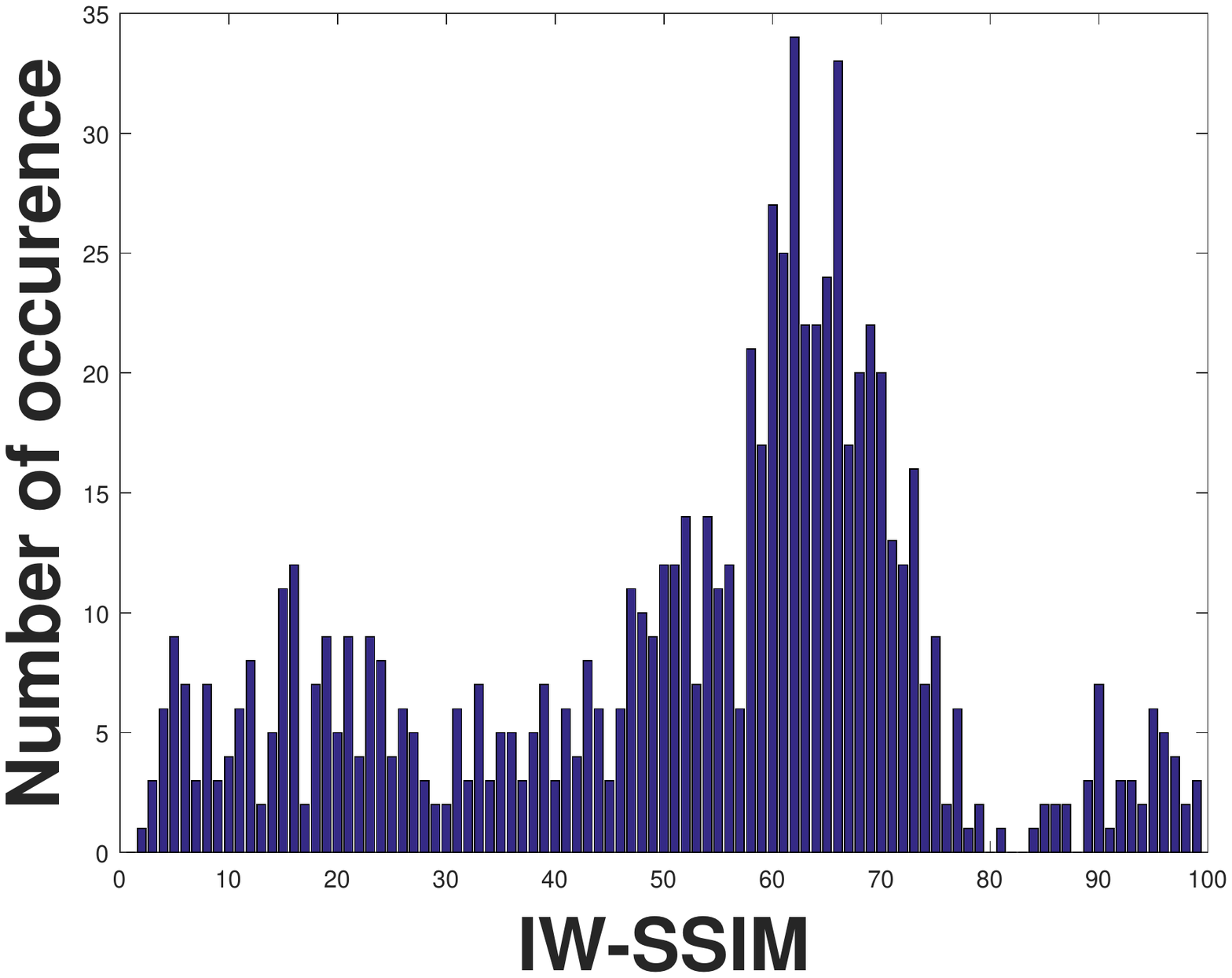}
  \vspace{0.03cm}
  \centerline{\footnotesize{(g)LIVE-IW-SSIM}}
        \vspace{-0.20cm}
\end{minipage}
  \vspace{0.20cm}
\hfill
\begin{minipage}[b]{0.28\linewidth}
  \centering
\includegraphics[width=0.8\linewidth, trim= 25mm 80mm 25mm 80mm]{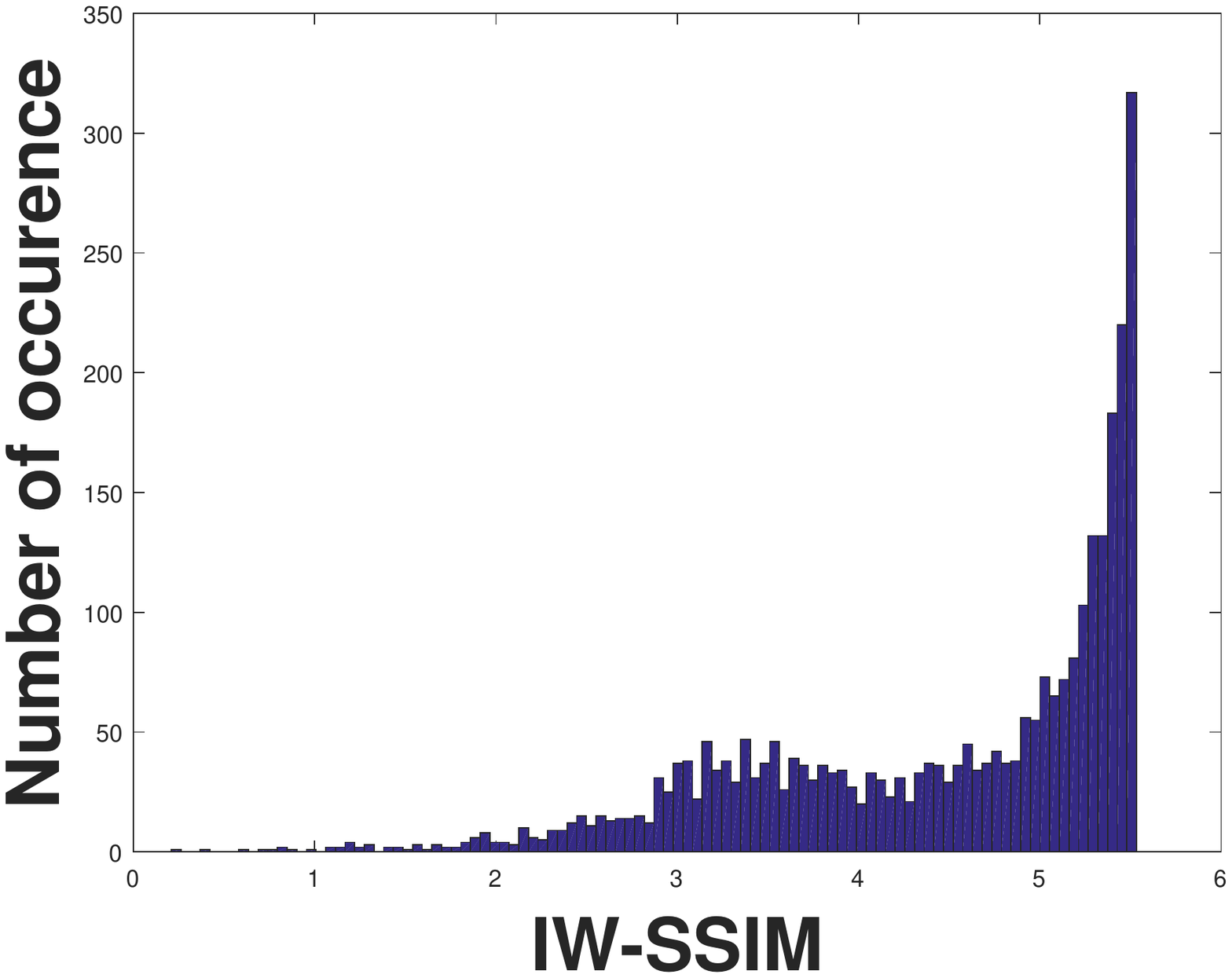}
  \vspace{0.03cm}
  \centerline{\footnotesize{(h) TID-IW-SSIM }}
        \vspace{-0.20cm}
\end{minipage}
  \vspace{0.20cm}
\hfill
\begin{minipage}[b]{0.28\linewidth}
  \centering
\includegraphics[width=0.8\linewidth, trim= 25mm 80mm 25mm 80mm]{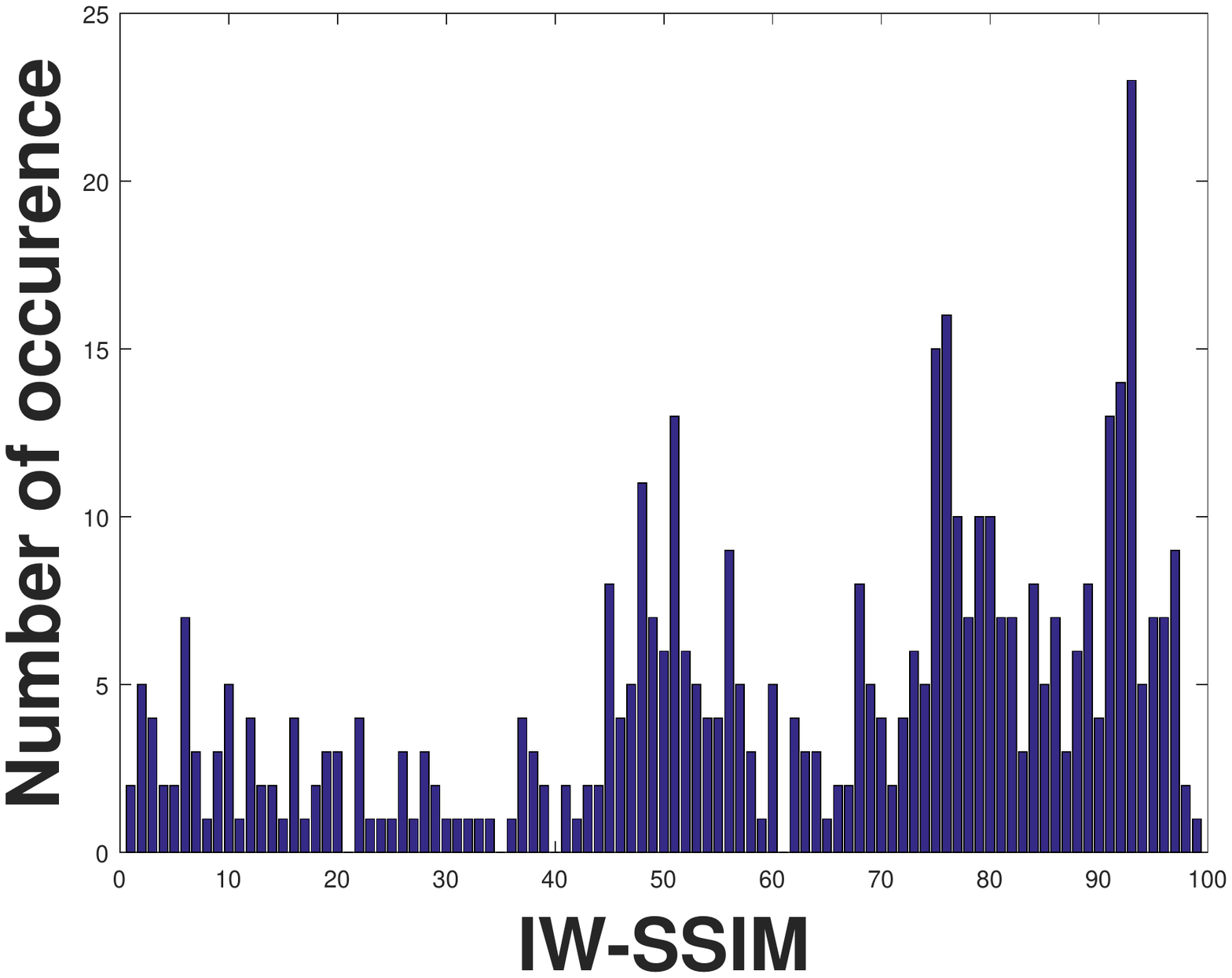}
  \vspace{0.03cm}
  \centerline{\footnotesize{(i) MULTI-IW-SSIM }}
  \vspace{-0.20cm}
\end{minipage}

\begin{minipage}[b]{0.28\linewidth}
  \centering
\includegraphics[width=0.8\linewidth, trim= 25mm 80mm 25mm 80mm]{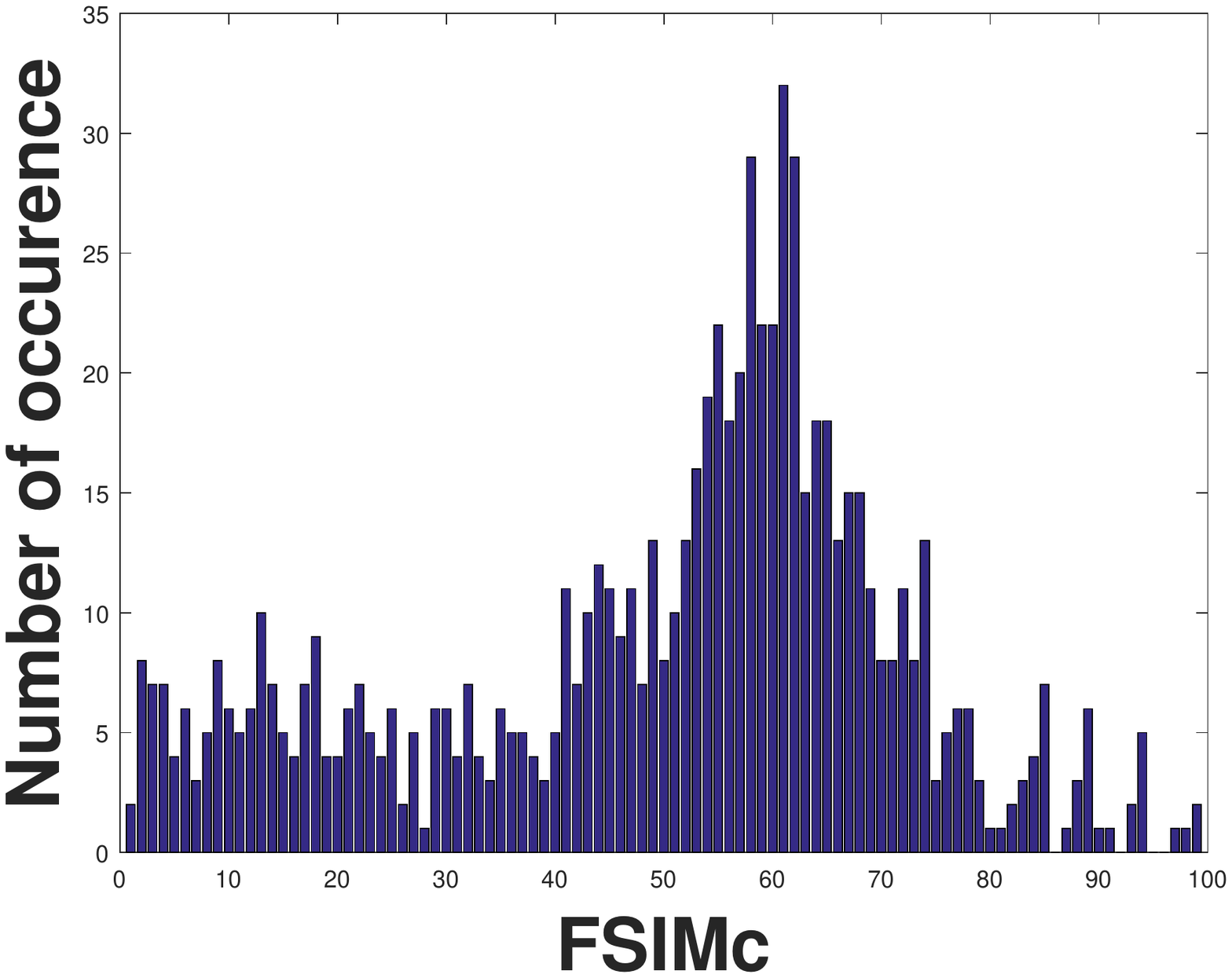}
  \vspace{0.03cm}
  \centerline{\footnotesize{(j)LIVE-FSIMc}}
  \vspace{-0.20cm}  
  \end{minipage}
  \vspace{0.20cm}
\hfill
\begin{minipage}[b]{0.28\linewidth}
  \centering
\includegraphics[width=0.8\linewidth, trim= 25mm 80mm 25mm 80mm]{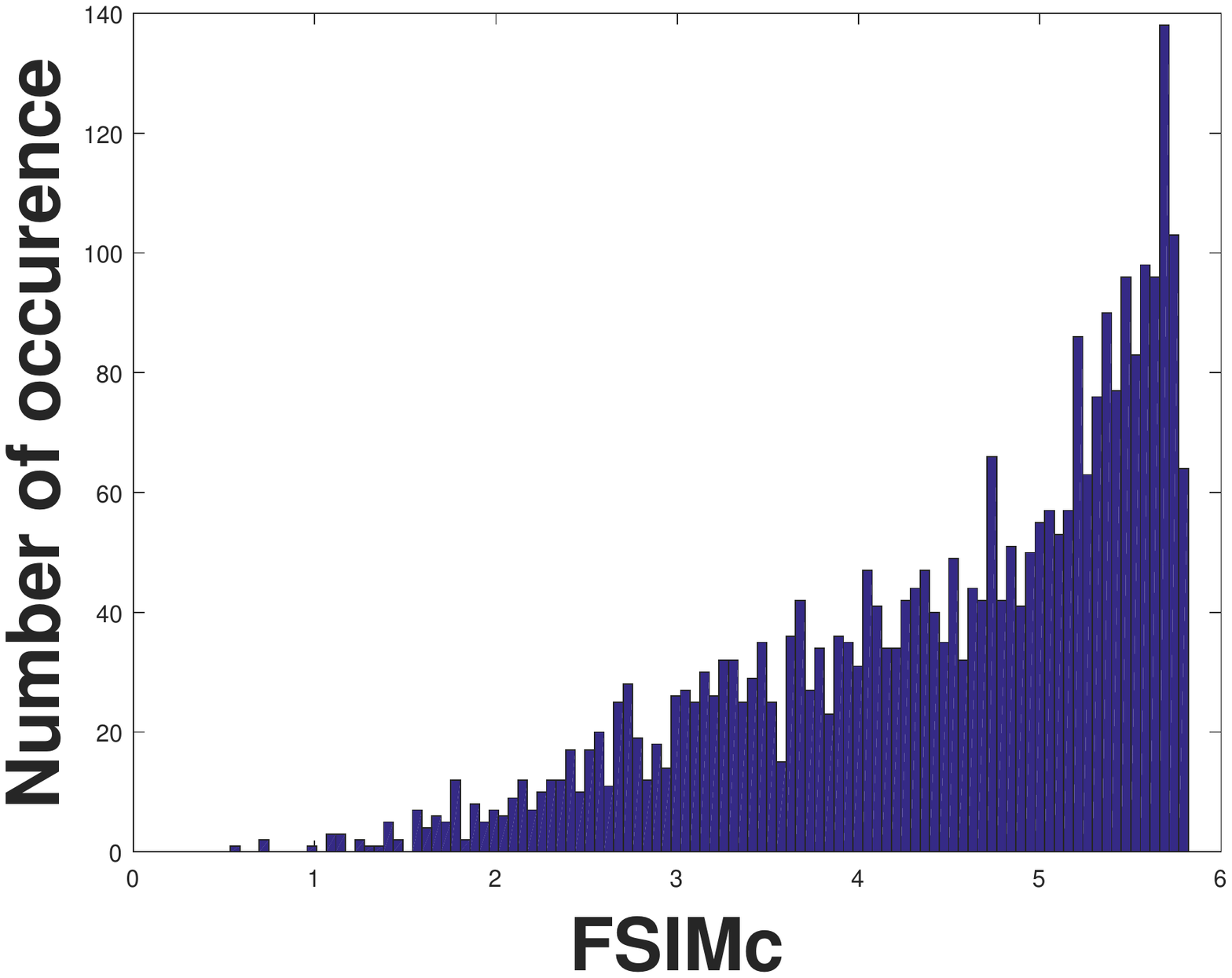}
  \vspace{0.03cm}
  \centerline{\footnotesize{(k) TID-FSIMc }}
  \vspace{-0.20cm}
\end{minipage}
  \vspace{0.20cm}
\hfill
\begin{minipage}[b]{0.28\linewidth}
  \centering
\includegraphics[width=0.8\linewidth, trim= 25mm 80mm 25mm 80mm]{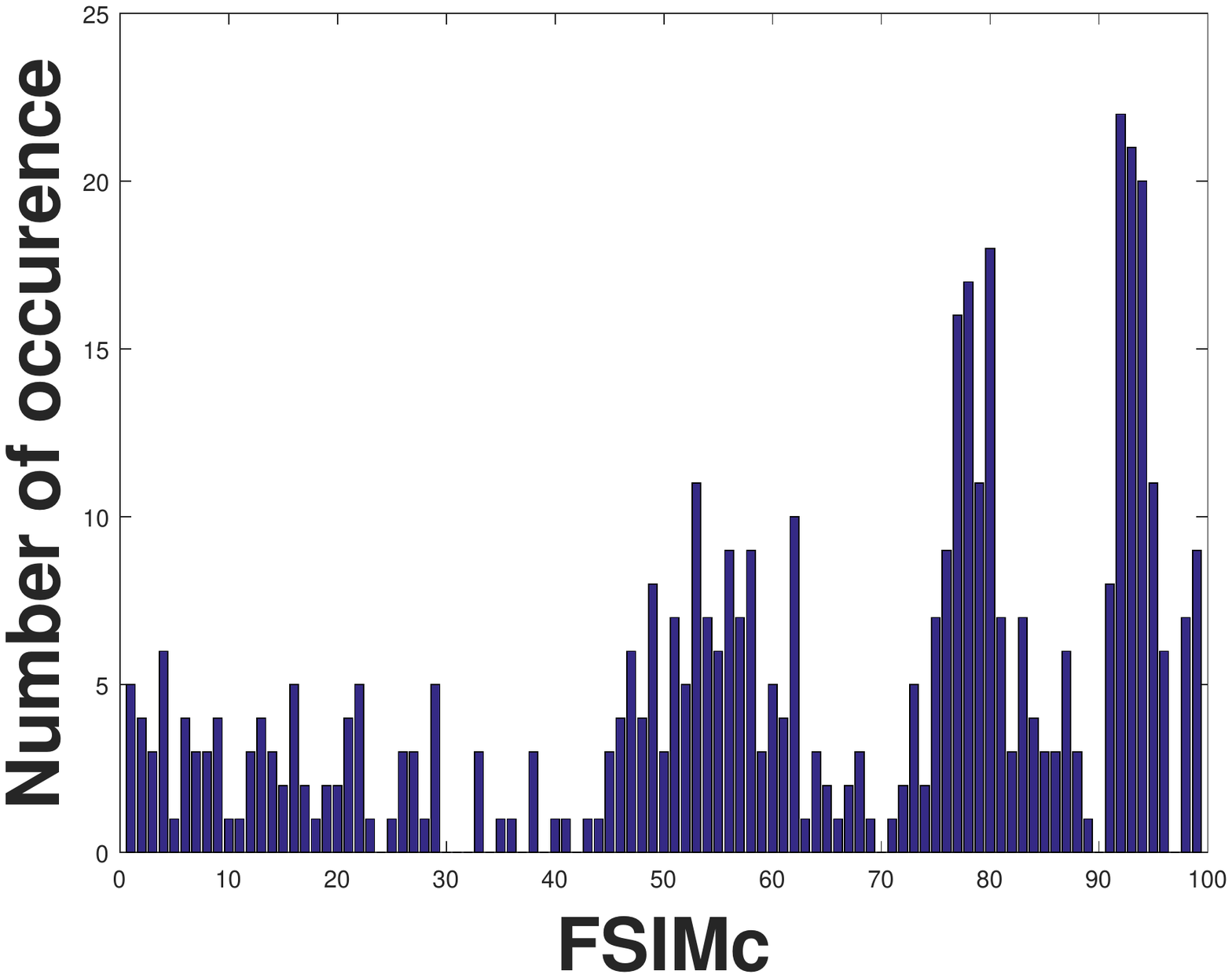}
  \vspace{0.03cm}
  \centerline{\footnotesize{(l) MULTI-FSIMc }}
  \vspace{-0.20cm}

\end{minipage}

\begin{minipage}[b]{0.28\linewidth}
  \centering
\includegraphics[width=0.8\linewidth, trim= 25mm 80mm 25mm 80mm]{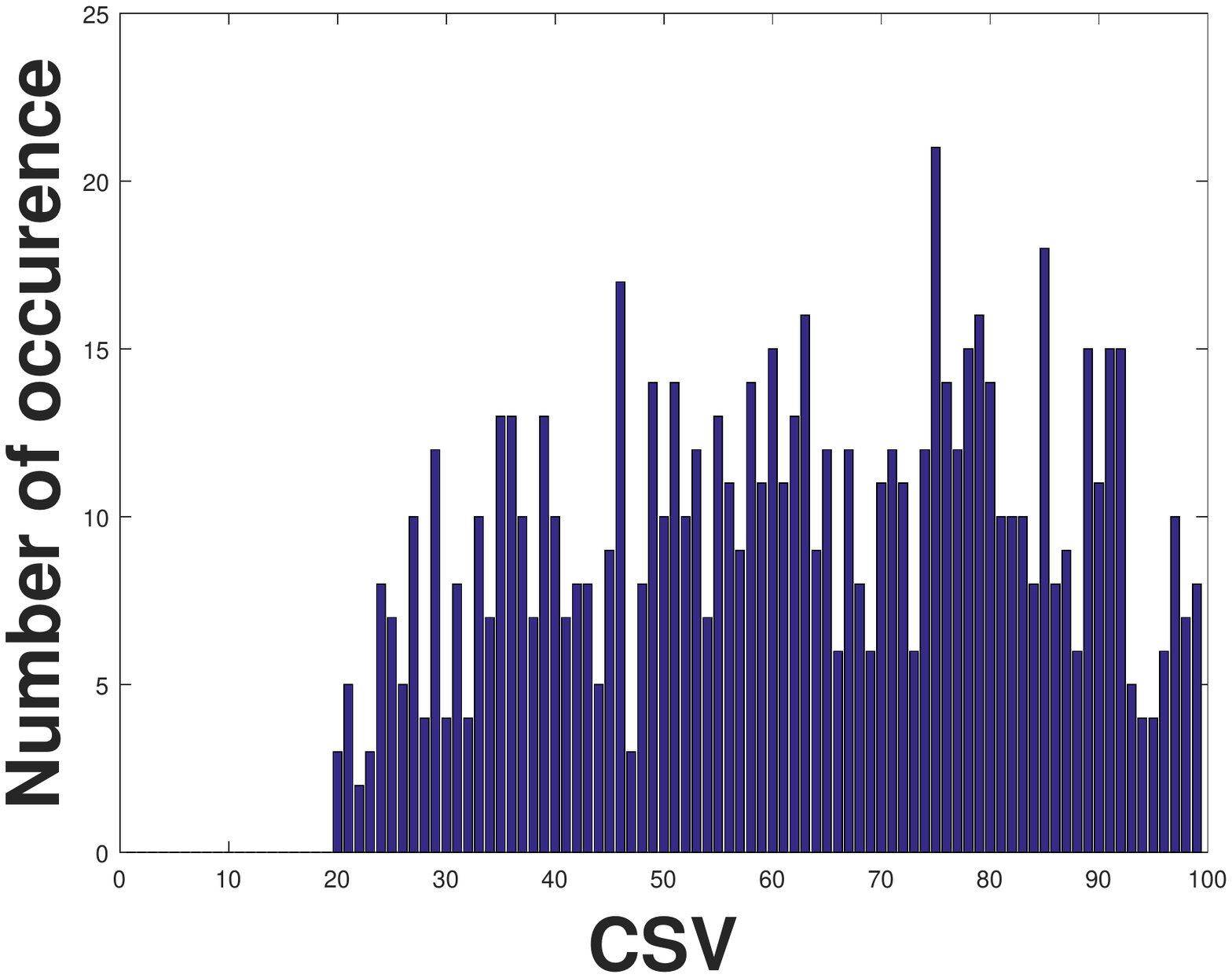}
  \vspace{0.03cm}
  \centerline{\footnotesize{(m)LIVE-CSV}}
  \vspace{-0.20cm}
\end{minipage}
  \vspace{0.20cm}
\hfill
\begin{minipage}[b]{0.28\linewidth}
  \centering
\includegraphics[width=0.8\linewidth, trim= 25mm 80mm 25mm 80mm]{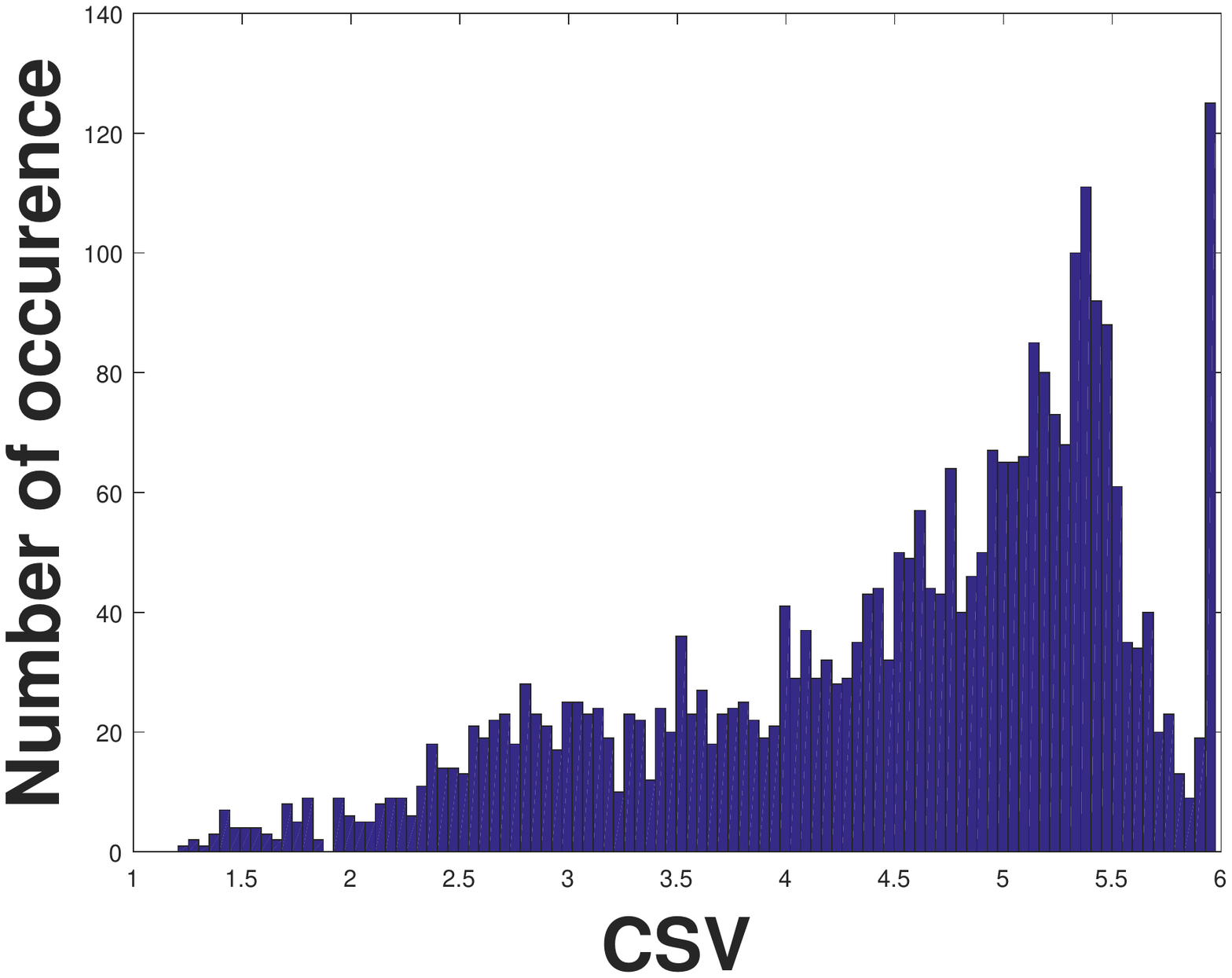}
  \vspace{0.03cm}
  \centerline{\footnotesize{(n) TID-CSV }}
  \vspace{-0.20cm}
\end{minipage}
  \vspace{0.20cm}
\hfill
\begin{minipage}[b]{0.28\linewidth}
  \centering
\includegraphics[width=0.8\linewidth, trim= 25mm 80mm 25mm 80mm]{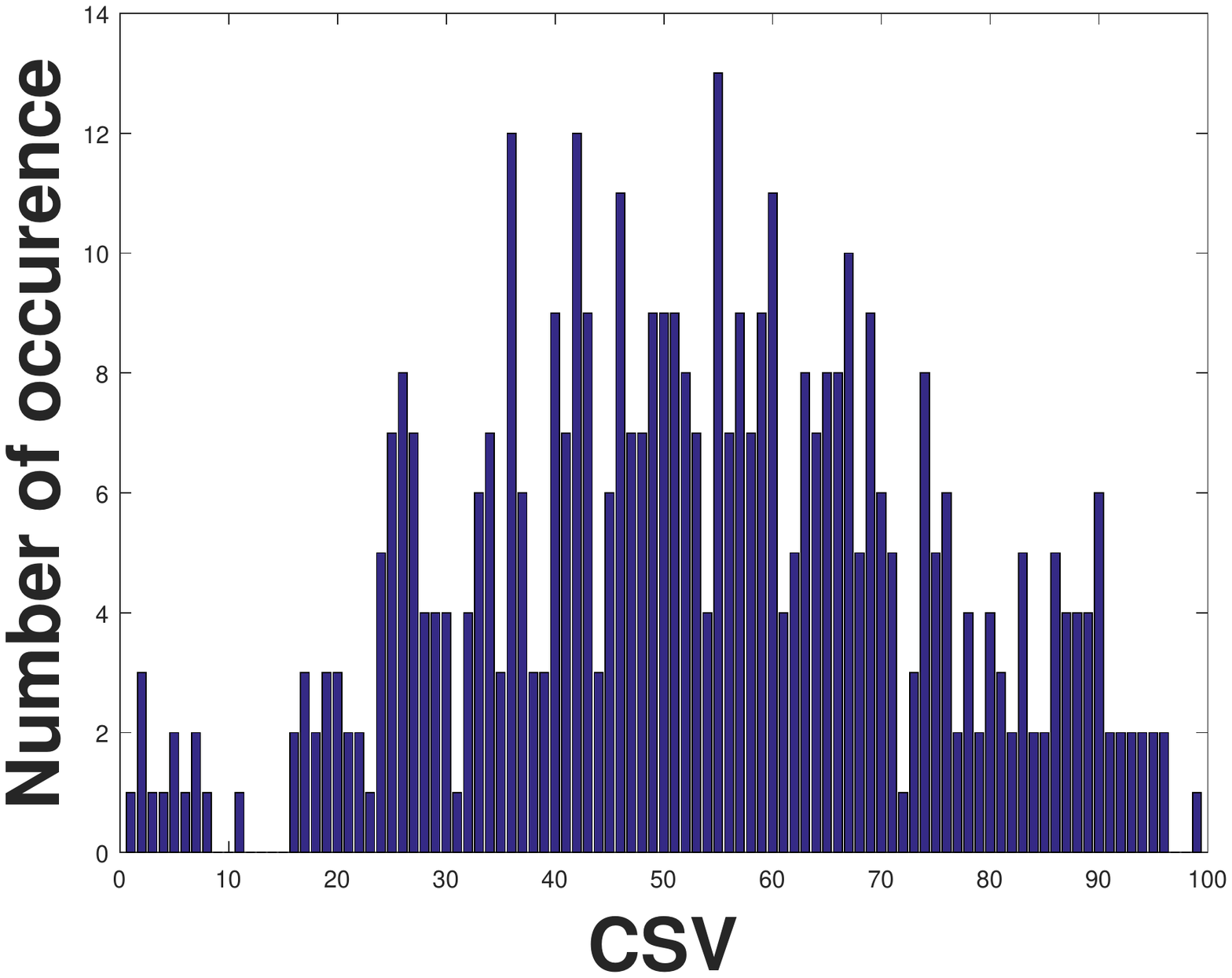}
  \vspace{0.03cm}
  \centerline{\footnotesize{(o) MULTI-CSV }}
  \vspace{-0.20cm}  
\end{minipage}

\caption{Histograms of the objective quality estimates and the subjective scores.}
\label{fig:Histograms}
\vspace{-7.0mm}
\end{figure}

\end{center}

In the LIVE database, the linearity of SSIM, IW-SSIM, and FSIMc are not very consistent throughout the quality range compared to \texttt{CSV}. In the TID database, the linearity of \texttt{CSV} is relatively more consistent than the linearity of other methods but the numerical range of the quality estimates in \texttt{CSV} is very limited. In the MULTI database, SSIM, IW-SSIM, and FSIMc follow a similar pattern, in which the quality estimates decrease monotonically but not too linearly. The distribution of \texttt{CSV} estimates are mostly monotonically decreasing and they have a higher linearity compared to other methods but estimates are bounded to a limited range. To measure the differences between distributions, we plot the histograms of the subjective scores and the regressed quality estimates in Fig. \ref{fig:Histograms}. The difference between the distributions are calculated using the common histogram distance metrics including Earth Mover's Distance (EMD), Kullback-Leibler (KL) divergence,  Jensen-Shannon (JS) divergence, histogram intersection (HI), and L2 norm. We provide the histogram distances in Table \ref{tab:hist_dist}. \texttt{CSV} has the minimum difference between the subjective scores and the quality estimates in terms of all the histogram difference measures. \texttt{CSV} is followed by FSIMc in the LIVE and the TID databases. In the MULTI database, \texttt{CSV} is followed by SSIM and IW-SSIM.  

\begin{center}

\begin{table*}[htbp!]
\vspace{-5.0mm}

\begin{adjustwidth}{-1.8cm}{}

  \centering
      \tiny
        \caption{Distributional difference between the subjective scores and the objective quality estimates.}

    \begin{tabular}{|c||c|c|c|c|c||c|c|c|c|c||c|c|c|c|c|}   \hline
    \multirow{2}[4]{*}{\textbf{Metric}} 
    
    & \multicolumn{5}{c||}{\textbf{Difference-LIVE}} & \multicolumn{5}{c||}{\textbf{Difference-TID}} & \multicolumn{5}{c|}{\textbf{Difference-Multi}} \\ \cline{2-16}     
    
          & \textbf{EMD}& \textbf{KL}& \textbf{JS}& \textbf{HI}& \textbf{L2}& \textbf{EMD}& \textbf{KL}& \textbf{JS}& \textbf{HI}& \textbf{L2}  & \textbf{EMD}& \textbf{KL}& \textbf{JS}& \textbf{HI}& \textbf{L2}    
          \\ \hline 

				\textbf{CSV} &\textbf{0.19} &\textbf{0.18} &\textbf{0.03} &\textbf{0.19} &\textbf{0.05}&\textbf{0.30}&\textbf{0.48}&\textbf{0.08}&\textbf{0.30}&\textbf{0.09} &\textbf{0.26}&\textbf{0.15}&\textbf{0.03}&\textbf{0.26}&\textbf{0.07} \\ \hline

	\textbf{FSIMc} &0.27 &0.29 &0.06 &0.27 &0.07&0.39&0.96&0.12&0.39&0.11 &0.42&0.51&0.10&0.42&0.12  \\

	\textbf{IW-SSIM} &0.30 &0.34 &0.07 &0.30 &0.07&0.50&1.67&0.19&0.50&0.18 &0.38&0.37&0.07&0.38&0.10  \\

	\textbf{SSIM} &0.28 &0.30 &0.06 &0.28 &0.07 &0.64&1.98&0.26&0.64&0.18 &0.38&0.41&0.08&0.38&0.10 \\ \hline

    \end{tabular}%
  \label{tab:hist_dist}
  \end{adjustwidth}{}

\end{table*}

\end{center}

\vspace{-2.0mm}
\section{Conclusion}
\label{sec:conclusion}
We propose a full-reference quality estimator based on color, structure, and visual system. Color difference and distance between color name descriptors are used to capture color-based degradations, a Laplacian of Gaussian operator is used to partially model contrast sensitivity of retinal ganglion cells, and a local normalization operator is used to mimic suppression mechanisms in cortical neurons. Primitive models of the contrast sensitivity and the suppression mechanisms along with the perceptual distance among color name descriptors are far away from being a comprehensive perceptual quality estimator. However, the performance of \texttt{CSV} articulates the importance of  combining various perception mechanisms in a single quality estimator. \texttt{CSV} is always among the top two performing objective quality estimators in terms of linearity, ranking or monotonic behavior in the LIVE, the MULTI, and the TID databases. To enhance the performance of \texttt{CSV} and formulate visual system characteristics more accurately, we need to extend the color difference equations and the color name distances, which are pixel-wise methods. Because the perception of a stimulus is not solely affected by the center but also the surround. Therefore, \texttt{CSV} can be extended with center-surround models to mimic perception mechanisms in a human visual system. Moreover, the feature maps in \texttt{CSV} can be obtained in multiple resolutions and fused to partially mimic the hierarchical nature of a human visual system. In terms of implementation, sequential processes in \texttt{CSV} including EMD and CIEDE2000 computations need to be parallelized to obtain a near-real time or real time image quality assessment algorithm.

\section*{References}


\end{document}